\documentclass{article}

\PassOptionsToPackage{numbers, compress}{natbib}
\PassOptionsToPackage{table}{xcolor}
\usepackage[preprint]{neurips_2026}

\usepackage[utf8]{inputenc}
\usepackage[T1]{fontenc}
\usepackage[hidelinks,hypertexnames=false]{hyperref}
\hypersetup{pdfborder={0 0 0}}
\usepackage{url}
\usepackage{booktabs}
\usepackage{amsfonts}
\usepackage{amsmath}
\usepackage{amssymb}
\usepackage{nicefrac}
\usepackage{microtype}
\usepackage{xcolor}
\usepackage{graphicx}
\usepackage{multirow}
\usepackage{array}
\usepackage{makecell}
\usepackage{enumitem}
\usepackage{colortbl}
\usepackage{bm}
\usepackage{listings}
\usepackage{float}
\usepackage{placeins}
\usepackage{needspace}

\graphicspath{{images/}} 

\definecolor{promptbg}{RGB}{245,245,245}
\definecolor{promptframe}{RGB}{210,210,210}
\definecolor{promptnum}{RGB}{150,150,150}
\definecolor{mygray}{gray}{0.9}

\lstdefinestyle{promptstyle}{
    backgroundcolor=\color{promptbg},
    basicstyle=\ttfamily\scriptsize,
    columns=fullflexible,
    breakatwhitespace=false,
    breaklines=true,
    breakindent=0pt,
    breakautoindent=false,
    captionpos=b,
    keepspaces=true,
    numbers=left,
    numbersep=8pt,
    numberstyle=\tiny\color{promptnum},
    showspaces=false,
    showstringspaces=false,
    showtabs=false,
    tabsize=2,
    frame=single,
    rulecolor=\color{promptframe},
    aboveskip=1em,
    belowskip=1em,
    xleftmargin=2.5em,
    framexleftmargin=2em,
    moredelim=**[l][\bfseries]{\#\#}
}

\AtBeginDocument{%
  }

\title{Bridging the Intention-Expression Gap: Aligning Multi-Dimensional Preferences via Hierarchical Relevance Feedback in Text-to-Image Diffusion}

\author{%
  Wenxi Wang \\
  Tongji University \\
  \texttt{2410915@tongji.edu.cn}
  \And
  Hongbin Liu \\
  Tongji University \\
  \texttt{2253689@tongji.edu.cn}
  \And
  Mingqian Li \\
  Tongji University \\
  \texttt{2252890@tongji.edu.cn}
  \AND
  Junyan Yuan \\
  Tongji University \\
  \texttt{Crispyyuan@tongji.edu.cn}
  \And
  Junqi Zhang \\
  Tongji University \\
  \texttt{zhangjunqi@tongji.edu.cn}
}

\begin{document}

\maketitle

\begin{abstract}
Users often possess a clear visual intent yet struggle to articulate it precisely in language~\cite{datta2008image}. This intention-expression gap renders the alignment of generated images with latent visual preferences a fundamental challenge in text-to-image diffusion models. Existing methods either require model training, sacrificing model-agnostic flexibility, or rely on textual feedback, which imposes a heavy cognitive burden by forcing users to articulate what they inherently cannot.
Although recent training-free methods utilize click-based binary preference feedback to reduce user effort, they force Foundation Models (FMs) to infer preferences at the semantic level.
When faced with inherently multi-dimensional preferences, FMs suffer from inference overload and fail to identify the user's exact preferred feature values under conflicting user signals.
Consequently, a flexible framework capable of achieving multi-dimensional feature alignment under the intention-expression gap remains absent.
To address this, we propose a Hierarchical Relevance Feedback-Driven (HRFD) framework. Recognizing that multiple preference features struggle to converge simultaneously, HRFD organizes them into a three-tier hierarchy and adapts the relevance feedback to enforce coarse-to-fine convergence, minimizing cognitive load. To bypass FM inference overload, HRFD decouples the overall process into multiple single-feature preference inference tasks, evaluating each feature independently. Furthermore, to overcome FMs' failure in identifying these preferred values, HRFD employs statistical mathematical inference methods to quantify the distribution divergence of features between "liked" and "disliked" image sets, achieving a more robust and transparent preference measurement that FMs fundamentally lack.
Crucially, HRFD operates entirely within the external text space, remaining strictly training-free and model-agnostic. Extensive experiments demonstrate that HRFD effectively captures the user's true visual intent, significantly outperforming baseline approaches in multi-dimensional preference alignment.

\end{abstract}

\section{Introduction}
\label{sec:intro}

Text-to-image generation based on diffusion models has emerged as a transformative technology in generative AI (e.g., Stable Diffusion \cite{podell2023sdxl,rombach2022high}, DALL·E2 \cite{dalle2}, and Midjourney). However, in real-world scenarios, users often possess a clear visual intent but struggle to express it precisely in language \cite{datta2008image}, inherently resulting in ambiguous prompts that generate images misaligned with their preferences. Next, we briefly review existing methods for preference alignment in diffusion models, which are categorized into training-based and inference-based routes, and subsequently present our contributions.

Early training-based methods (e.g., DDPO \cite{DDPO}, ImageReward \cite{xu2023imagereward}, Diffusion-DPO \cite{Diffusion-DPO}, IterComp \cite{IterComp}) attempt to achieve preference alignment by training reward models on extensive human annotations. However, this non-interactive paradigm fundamentally fails when ambiguous prompts obscure the true intent. To mitigate this, interactive training methods leverage user dialogue and model fine-tuning (e.g., Twin-Co \cite{Twin-Co}) or train foundation model agents on multi-turn preference datasets (e.g., PASTA \cite{Pasta}). While effective, they either impose heavy cognitive loads by relying on explicit textual dialogue or incur prohibitive training costs. Conversely, inference-based approaches resolve heavy training costs. Early inference methods guide generation by manipulating the latent space via automated evaluators (e.g., DNO \cite{DNO}, DAS \cite{DAS}, DEMON \cite{Demon}), while recent approaches incorporate FMs and external knowledge bases for autonomous error correction (e.g., SIDiffAgent \cite{SIDiffAgent}, RAVEL \cite{RAVEL}, RAISE \cite{RAISE}). Nevertheless, their complete absence of user interaction prevents alignment under semantically ambiguous prompts.
To address the lack of interaction, subsequent approaches introduce explicit user interventions. Some assist manual prompt refinement through advanced interfaces like PromptCharm \cite{PromptCharm} and PrompTHis \cite{PrompTHis}. Others leverage FMs to iteratively refine prompts driven by detailed textual critiques, as seen in DSPy \cite{DSPy}, ConstitutionMaker \cite{ConstitutionMaker}, and Promptimizer \cite{Promptimizer}. However, these paradigms impose a significant cognitive load by forcing users to articulate what they inherently cannot. Following the trend toward lower cognitive load, the current state-of-the-art (SOTA) method, APPO \cite{APPO}, replaces tedious textual dialogue with click-based binary preference selections. By leveraging FMs to infer user preferences from multi-round interaction histories, APPO significantly reduces the cognitive burden while achieving superior alignment performance compared to previous methods.
Despite its SOTA status, APPO exposes fundamental limitations when applied to multi-dimensional feature alignment. A user typically possesses multiple preference features concurrently, such as artistic style, color tone, and composition. Because APPO relies on FMs to infer preferences strictly at a holistic semantic level across multiple features, it faces two critical bottlenecks. First, FMs suffer from severe inference overload when attempting to simultaneously infer and balance preferences across these multiple dimensions. Second, this holistic paradigm is highly vulnerable to conflicting user signals. For instance, if a user selects Image A for its composition and Image B for its color tone, but Image B features a composition that directly opposes Image A, it introduces contradictory semantic signals. This confounds the FM, preventing it from identifying the exact preferred feature values under conflicting signals. Therefore, there is an urgent need for a flexible framework capable of achieving multi-dimensional feature alignment under the intention-expression gap.

To address this issue, this work aims to make the following novel contributions:
\begin{itemize}[noitemsep, topsep=0pt,leftmargin=1em]
    \item We propose the Hierarchical Relevance Feedback-Driven (HRFD) framework for multi-dimensional preference alignment. HRFD organizes features into a three-tier hierarchy and adapts the relevance feedback mechanism to enforce coarse-to-fine convergence via lightweight binary signals, effectively addressing the challenge that multiple preference features cannot converge simultaneously.
    \item We design a mathematically grounded preference inference pipeline supported by an expert-curated feature repository. By decoupling complex visual intent into independent feature-level tasks and employing statistical measures to quantify preference strengths, it avoids the inference overload typical of FM-driven simultaneous reasoning. Simultaneously, it enables a more robust, transparent preference measurement that FMs fundamentally lack.
\item Experimental validation demonstrates that HRFD achieves superior performance over the baseline approach, with ablation studies confirming the necessity of each core component.

\end{itemize}

\section{Related Work}
\label{sec:related work}

\subsection{Text-to-Image Diffusion Models}
\label{sec:text-to-image}
Diffusion models were first proposed by Sohl-Dickstein et al.~\cite{sohl2015deep},
which generate data by learning the reverse denoising process. In the domain of
text-to-image generation, research has primarily focused on two challenges. The
first centers on enhancing foundational visual quality and ensuring semantic
consistency between generated images and text prompts. Three technical routes have
been developed: dataset quality improvement, which filters and constructs
high-quality training corpora to provide better supervision signals~\cite{betker2023improving,
dai2023emu, segalis2023picture, sun2025dreamsync, an2025agfsync}; model
architectural optimization, which refines cross-attention mechanisms and attention
allocation to strengthen the model's adherence to textual prompts~\cite{feng2022training,
hertz2022prompt, chefer2023attend}; and training strategy optimization, which has
advanced from supervised fine-tuning to AI-driven, fully automated pipelines~\cite{sun2025dreamsync, 
an2025agfsync}. The second challenge is achieving alignment with user
preferences. As detailed in Section~\ref{sec:intro}, the field still lacks an
interactive framework capable of achieving multi-dimensional feature alignment
under the intention-expression gap.

\subsection{Relevance Feedback}
\label{sec:relevance}
Relevance Feedback (RF) is a fundamental technique in Information Retrieval, addressing the deviation between the initial query and the latent intent in user retrieval tasks. Rocchio et al.~\cite{rocchio1971relevance} first established the benchmark for this field in text retrieval. Subsequently, to adapt to the characteristics of Content-Based Image Retrieval, studies extended this paradigm to the visual modality, capturing user feedback through linear weighting and probabilistic models~\cite{rui1998relevance, zhou2003relevance}. To break through linear constraints and improve efficiency in high-dimensional spaces, researchers introduced Support Vector Machines and nonlinear classification boundaries, successfully modeling users' nonlinear preferences~\cite{tao2006asymmetric, tong2001support, hong2000incorporate}. Based on this, to address the challenge posed by high-dimensional indexing, Zhang~\cite{zhang2005using} utilized high-dimensional indexing techniques to accelerate the efficiency of feedback retrieval. Furthermore, to enhance global exploration capabilities and improve global convergence efficiency in complex feature spaces, works such as~\cite{broilo2010stochastic, kanimozhi2015integrated, mahmood2022hybrid} adopted swarm intelligence algorithms, including Particle Swarm Optimization and the Firefly Algorithm, enhancing retrieval efficiency. With the development of deep learning and FMs, RF has begun to evolve towards conversational retrieval and query expansion. Some works utilized reinforcement learning to optimize long-term interaction rewards~\cite{zou2019reinforcement}, while others combined Large Language Models to construct semantic collaboration layers~\cite{mackie2023generative, sharma2025can}, significantly improving query expansion capabilities in complex scenarios. Despite the fact that existing relevance feedback is quite mature in retrieval tasks, it remains rarely explored in generative tasks.

\section{Method}
\label{sec:method}

\begin{figure}[!htbp]
  \centering
  \includegraphics[width=\linewidth]{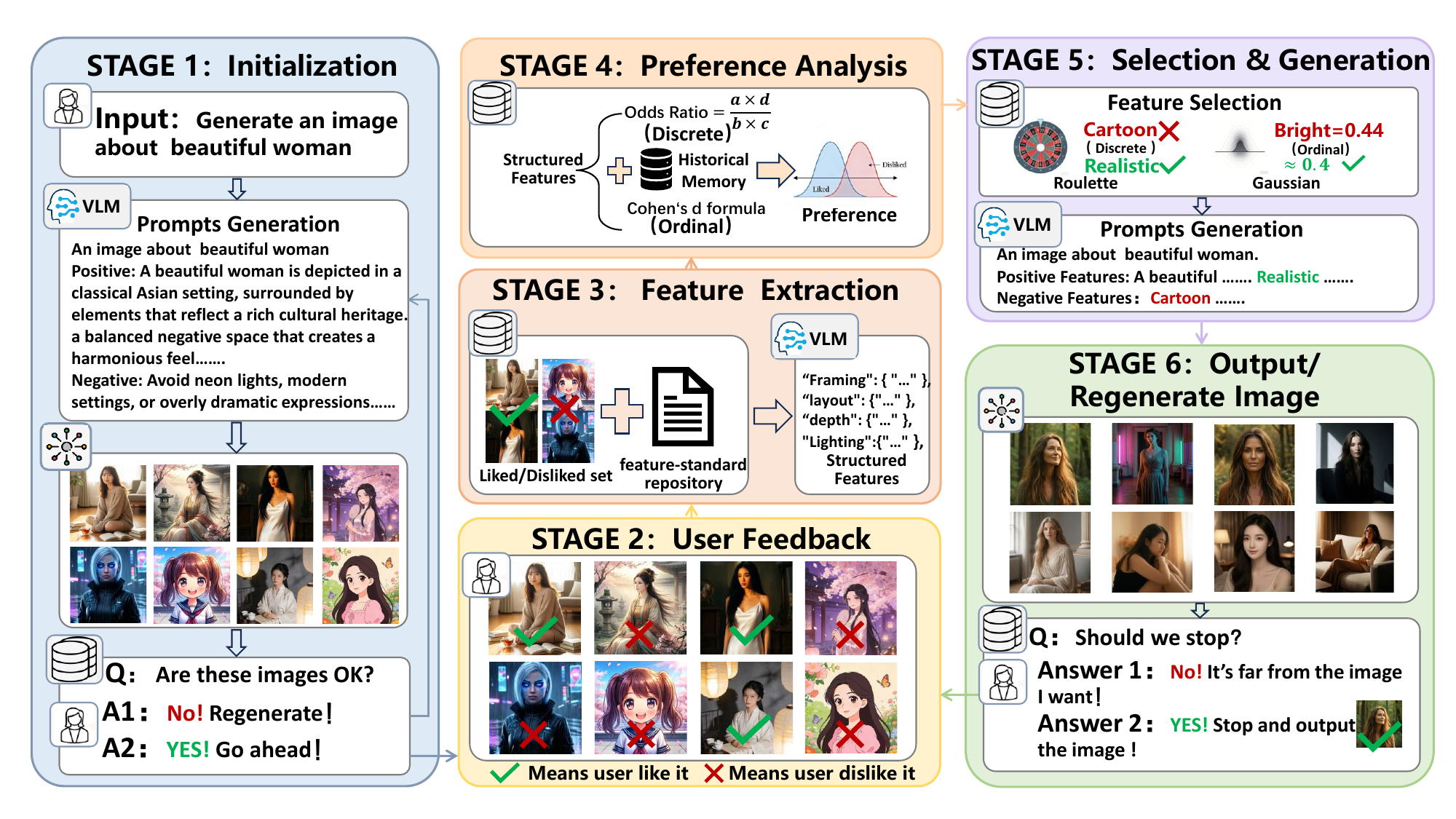}
  \vspace{-15pt} % <<--- 榨取空间的神器，数值可以根据需要微调（比如 -8pt 或 -12pt）
  \caption{The overall framework of the proposed HRFD method.}
  \label{fig:pic1}
\end{figure}

This section presents a Hierarchical Relevance Feedback-Driven (HRFD) interactive framework for diffusion-based image generation, as illustrated in Figure~\ref{fig:pic1}. Specifically, during initialization, HRFD generates diverse candidate images from the user's base prompt, which the user annotates with binary "like/dislike" feedback. HRFD extracts multi-dimensional image features using Vision-Language Models (VLMs) based on an expert-curated feature repository. Then, an information-theoretic weighted cumulative preference analysis is employed to infer user preferences from the feedback. Finally, preference information is converted into natural language prompts that drive the diffusion model to generate new candidate images for user annotation. Through multiple interaction rounds, HRFD progressively refines its understanding of user intent, enabling generated results to converge toward the desired visual outcome. The specific methods to accelerate this iterative convergence are detailed in Appendix \ref{app:Interactive_procedure}.

\subsection{Multi-Dimensional Feature Extraction}
\label{sec:feature_extraction}

In HRFD, after users express visual preferences through binary ``like/dislike'' annotations, the core challenge is transforming these abstract preferences into concrete directions that guide generative models. Current methods typically rely on FMs to perform preference inference by directly utilizing the image-generation prompt alongside per-image user feedback within an open, unbounded semantic space. However, this paradigm suffers from two fundamental limitations. First, the prompt captures only the user's explicit semantic intent, inevitably omitting the rich visual attributes manifested in the images (such as compositional balance), leading to incomplete preference characterization. Second, conducting inference within an open, unbounded semantic space lacks explicit structural boundaries, which renders the inference process overly broad and makes it difficult to pinpoint specific aesthetic details.
To address these limitations, HRFD performs structured feature extraction on images based on an expert-curated feature-standard repository. This ensures that key visual attributes of images are comprehensively captured without omission, and constrains preference inference within a well-defined feature space, transforming the otherwise unbounded inference problem into a structured, targeted analysis over precise aesthetic dimensions.

\textbf{Feature Standard Repository Construction.}
The feature-standard repository in this work is constructed based on expert knowledge from~\cite{kong2016photo}, and is further supplemented and refined in collaboration with five senior photographers. The repository is decomposed into three hierarchical levels: Global Attributes (e.g., artistic style), Spatial and Chromatic Structure (e.g., composition), and Fine-grained Details (e.g., texture). The detailed design rationale for this hierarchy is provided in Appendix \ref{app:Design Rationale}.
Crucially, users interact only with the coarse semantic meaning of each tier, the detailed features within each tier and their convergence are handled transparently by HRFD, imposing no additional cognitive burden. The statistical properties of these features are categorized into discrete features (lacking inherent ordering, e.g., style) and ordinal features (possessing explicit ordering, e.g., brightness).
The complete feature-standard repository is detailed in Appendix \ref{app:appendix_features}.

\textbf{Structured Feature Extraction.}
The above feature-standard repository defines ``which features should be extracted''. Next, we guide VLMs to perform multi-dimensional structured feature extraction using a prompt with three core components. First, a role constraint designates the model as an 'expert image analyst' to ensure objective evaluation. Second, to mitigate hallucinations, the extraction operates strictly as a constrained classification task, mapping visual content exclusively to our predefined repository categories. Third, a strict JSON-only format constraint renders the results directly parsable for downstream algorithmic processing.
The reliability of this VLM-based feature extraction has been empirically validated to ensure the mathematical soundness of our pipeline, with detailed evaluations and the extraction meta-prompt provided in Appendices \ref{app:feature_extraction} and \ref{app:stage1_metaprompt}, respectively.

\subsection{Information-Theoretic Weighted Cumulative Preference Analysis}
\label{sec:information-theoretic}
With the structured features extracted, we mathematically quantify user preferences. Given their distinct properties, discrete and ordinal features require tailored analysis strategies.

\textbf{Discrete Features Analysis.}
For discrete features, we focus on evaluating user preferences for each specific candidate value $v$ (e.g., \emph{art style} = ``realistic''). At each interaction round $t$, we construct a $2 \times 2$ contingency table to characterize the relationship between the presence of value $v$ and user feedback. Let $A_t$, $B_t$, $C_t$, and $D_t$ denote the sample counts at round $t$ for ``liked and $v$ is present,'' ``disliked and $v$ is present,'' ``liked and $v$ is absent,'' and ``disliked and $v$ is absent,'' respectively. Based on this, the Odds Ratio (OR) is employed to quantify the direction and strength of user preference for value $v$:
\begin{equation}
\text{OR}_t(v) = \frac{A_t \cdot D_t}{B_t \cdot C_t}
\end{equation}
where $\text{OR}_t(v) > 1$ indicates a positive preference for the feature value $v$ at round $t$, while $\text{OR}_t(v) < 1$ indicates a negative preference.

The term $\text{OR}_t(v)$ captures only the immediate association within round $t$. Relying solely on it would cause earlier preferences to be forgotten. Unlike FM-driven approaches that concatenate feedback into an ever-expanding prompt and inevitably suffer from context degradation, HRFD forms a continuous preference memory by incrementally accumulating single-round feedback into a lightweight mathematical state, ensuring robust reasoning regardless of interaction length. Furthermore, when fusing historical data, non-discriminative rounds (i.e., when $v$ appears in all or none of the samples) provide no useful information. To address this, we introduce an information-entropy-based adaptive weight. Let $e_t(v)$ denote the occurrence rate of value $v$ among candidate images at round $t$; the weight is defined as:
\begin{equation}
w_t(v) = -e_t(v) \log_2 e_t(v) - (1-e_t(v))\log_2(1-e_t(v))
\end{equation}
This weight $w_t(v)$ approaches zero as $e_t(v) \to 0$ or $e_t(v) \to 1$, and reaches its maximum when $e_t(v) = 0.5$, effectively filtering out non-informative rounds. Accordingly, the cumulative odds ratio for value $v$ is computed by aggregating weighted historical statistics:
\begin{equation}
\text{OR}(v) = \frac{\left(\displaystyle\sum_{t=1}^{T} w_t(v) \cdot A_t\right) \cdot \left(\displaystyle\sum_{t=1}^{T} w_t(v) \cdot D_t\right)}{\left(\displaystyle\sum_{t=1}^{T} w_t(v) \cdot B_t\right) \cdot \left(\displaystyle\sum_{t=1}^{T} w_t(v) \cdot C_t\right)}
\end{equation}
Consistent with the single-round metric, a cumulative $\text{OR}(v) > 1$ signifies a positive preference for value $v$ across the entire interaction history, whereas $\text{OR}(v) < 1$ implies the opposite.
To ensure numerical stability, the cumulative counts of $A$, $B$, $C$, and $D$ are each initialized to 1.

When resolving the aforementioned conflicting signals across multi-dimensional preferences, 
% FMs are probabilistic autoregressive models capable only of opaque semantic approximation. They struggle with exact feature tracking and mathematical counting over multi-turn interaction contexts. In contrast, 
our OR method provides a more statistically robust and transparent solution. By leveraging statistical disambiguation, it effectively distinguishes which feature values the user genuinely prefers from those that are merely noise, even under conflicting signals.

\textbf{Ordinal Features Analysis.}
\label{sec:ordinal_features}
For ordinal features (e.g., brightness) normalized to $[0,1]$, Cohen's $d$ is a natural choice for quantifying the ordinal features. However, applying Cohen's $d$ directly in the raw metric space fails under centralized preferences due to a \emph{cancellation effect}: when a user prefers a central value (e.g., 0.5) while rejecting both extremes (e.g., 0.1 and 0.9), the liked and disliked group means both converge toward $0.5$, driving the effect-size numerator to zero despite a clear preference signal.

To resolve this, we shift the analysis to an absolute deviation
space anchored at a \emph{cumulative preference center}
$c_{\text{pref}}(f)$, which is defined as the weighted mean of all
historically liked samples:
\begin{equation}
c_{\text{pref}}(f)
  = \frac{\displaystyle\sum_{t=1}^{T}\sum_{i \in \mathcal{S}_{\mathrm{liked},t}}
          w_t(f)\cdot v_i^t}
         {\displaystyle\sum_{t=1}^{T}\sum_{i \in \mathcal{S}_{\mathrm{liked},t}}
          w_t(f)},
\end{equation}
where $w_t(f)$ is a variance-based weight that down-weights rounds whose candidate images exhibit low feature variance and thus carry little discriminative signal (full derivation in Appendix~\ref{app:ordinal}). $c_{\text{pref}}(f)$ shifts dynamically with each round to track the user's evolving intent.

Within this absolute deviation space, we compute the weighted mean
deviation $\mu'_g(f)$ and pooled standard deviation
$\sigma'_{\mathrm{pooled}}(f)$ over $|v_i^t - c_{\text{pref}}(f)|$
for $g\in\{\mathrm{liked},\mathrm{disliked}\}$
(full formulations in Appendix~\ref{app:ordinal}), and
quantify preference intensity via Cohen's $d$:
\begin{equation}
d(f) = \frac{\mu'_{\mathrm{disliked}}(f) - \mu'_{\mathrm{liked}}(f)}
            {\sigma'_{\mathrm{pooled}}(f)}.
\end{equation}
A larger $d(f)$ reflects a more emphatic preference for the feature
level at $c_{\text{pref}}(f)$, and serves as the threshold criterion
for Gaussian sampling in Section~\ref{sec:gaussian_sampling}.

\subsection{Preference-Driven Feature Selection and Prompt Generation}
\label{sec:preference-driven}
Upon completing the cumulative preference analysis, HRFD selects
appropriate feature values and constructs natural-language prompts
for the next generation round via the following strategies.

\textbf{Roulette Wheel Sampling for Discrete Features.}
To balance exploration and exploitation, we employ roulette-wheel sampling based on cumulative ORs. Because the Odds Ratio is an asymmetric multiplicative metric, we project it into a symmetric Log-Odds space and apply the Softmax function. This simplifies to a direct proportional probability:

\begin{equation}
P(v_k \mid f) = \frac{\exp(\ln(\text{OR}(v_k)))}{\displaystyle\sum_{j=1}^{K} \exp(\ln(\text{OR}(v_j)))} = \frac{\text{OR}(v_k)}{\displaystyle\sum_{j=1}^{K} \text{OR}(v_j)}
\end{equation}

With this distribution rigorously established, a specific value for each feature is independently drawn via roulette-wheel selection. This sampling mechanism naturally transitions from broad exploration (when early feedback is sparse and ORs are uniform) to sharp exploitation as user intent converges.
% It is worth noting that sampling features independently may occasionally result in conflicting combinations. However, our framework effectively mitigates this issue in the subsequent stage. By treating these sampled values as soft constraints during prompt generation (Section 3.3.3), the VLM uses its pre-trained knowledge to naturally resolve any conflicts. This ensures the generated prompts remain logical and visually coherent.

\textbf{Gaussian Sampling for Ordinal Features.}
\label{sec:gaussian_sampling}
For ordinal features, we adopt an adaptive thresholding strategy guided by the cumulative effect size $d(f)$. 
Since the absolute deviation $|v_i - c_{\text{pref}}|$ is a direct transformation of the original metric, Cohen's $d$ retains its dimensionless property in this space. Consequently, we follow Cohen's established convention~\cite{cohen2013statistical} and directly adopt $0.8$ as a practical threshold to distinguish clear from ambiguous preference signals.

If $d(f) < 0.8$, indicating an ambiguous preference, we uniformly sample a continuous value $x_f \sim \mathcal{U}(0, 1)$ to ensure sufficient exploration. Conversely, a strong preference ($d(f) \ge 0.8$) triggers exploitation, where we sample from a truncated Gaussian distribution anchored at the preference center: $x_f \sim \mathcal{N}_{[0,1]}\!\left(c_{\text{pref}}(f), \sigma_{\text{liked}}^2(f)\right)$. To maintain a robust exploration range and prevent variance collapse under conflicting feedback, the sampling variance $\sigma_{\text{liked}}^2(f)$ is derived from historically liked samples in the original metric space (see Appendix~\ref{app:ordinal} for details). Finally, $x_f$ is mapped to the nearest valid ordinal level to determine the feature value, effectively balancing exploration and convergence.

\textbf{Feature-Driven Prompt Generation.}
Sampled feature values are translated into final prompts via a VLM, guided by a meta-prompt (Appendix~\ref{app:stage3_metaprompt}). The user's initial text strictly serves as the inviolable semantic core (\textbf{hard constraints}), whereas the sampled features act as \textbf{soft constraints}. Although foundation models inherently struggle with exact feature counting and reliable multi-step mathematical reasoning over lengthy contexts~\cite{liu2024lost,dziri2023faith}, extensive literature has demonstrated their exceptional capability in linguistic synthesis and prompt rewriting~\cite{ouyang2022training,hao2022optimizing}. 
Specifically, when some features are provided, the VLM seamlessly integrates them into a fluent textual description. Even when conflicting feature values are occasionally sampled, the model leverages its pre-trained language priors to resolve such semantic inconsistencies naturally, generating a grammatically coherent description without disrupting the core narrative. Finally, prompt lengths are strictly bounded to 30--80 words (\textbf{output requirements}) to maintain semantic focus before generating the next batch of candidate images.

\section{Experiments}
\label{sec:experiments}

\subsection{Experimental Setup}
\label{sec:experimental setup}
Experiments were conducted with 15 human participants who received standardized training prior to the study. To ensure feedback validity, participants adhered to four operational principles.

\begin{itemize}[noitemsep, topsep=0pt, leftmargin=1em]
\item \textbf{Subject Clarity:} The initial prompt must clearly specify the core visual subjects (single or multiple entities).
\item \textbf{Intent-Driven Selection:} Selections should prioritize images that best capture the user's holistic visual intent, avoiding choices driven solely by a single appealing detail. This ensures that HRFD receives robust feedback signals for comprehensive preference inference.
\item \textbf{Preference Consistency:} Users must maintain a stable preference direction without mid-session shifts, as HRFD is designed to converge on a specific latent intent.
\item \textbf{Hierarchical Awareness:} Users only need to grasp the coarse-to-fine progression of the three-tier feature hierarchy, with no requirement to understand any fine-grained details of the specific features within each tier.
\end{itemize}

% \begin{itemize}[noitemsep, topsep=0pt, leftmargin=1em]
% \item \textbf{Subject Clarity:} Participants were required to ensure that the core visual subjects (encompassing either single or multiple entities) were clearly specified in the initial prompt.
% \item \textbf{Intent-Driven Selection:} Participants were instructed to select images that best matched their overall visual intent across multiple dimensions, avoiding choices based solely on isolated local features. This ensures the HRFD receives robust feedback signals for multi-dimensional preference inference.
% \item \textbf{Preference Consistency:} Participants were required to maintain a stable and consistent preference direction throughout the interaction session. We instructed participants to avoid mid-session intent shifts or back-and-forth preference changes, as the HRFD is designed to converge on a specific latent intent rather than adapt to drifting preferences.
% \item \textbf{Hierarchical Awareness:} Participants were informed that the HRFD aligns preferences coarse-to-fine across three tiers: Global Attributes, Spatial and Chromatic Structure, and Fine-grained Details, requiring no explicit knowledge of the underlying features at each tier.
% \end{itemize}

The entire interaction is strictly click-based, requiring no textual feedback. Computations were performed on a 22GB RTX 2080 Ti GPU, utilizing the distilled FLUX.2-Klein for rapid image synthesis and the Llama-4-Maverick-17B API for feature extraction and prompt generation. Detailed hyperparameters and latency analysis are provided in the Appendix~\ref{app:computational_efficiency}.

\subsection{Baselines and Evaluation Metrics}
\label{baselines}
\textbf{Baselines.}
We compare HRFD against two methods. The first is APPO~\cite{APPO}, the state-of-the-art approach for click-only alignment. APPO concatenates multi-round interaction histories to drive a foundation model to infer holistic user preferences across multiple visual features.
The second baseline is LPO (LLM-based Prompt Optimization), an open-loop method that directly rewrites the initial ambiguous prompt without incorporating iterative feedback. We include LPO to demonstrate that powerful LLMs alone cannot bridge the intention--expression gap, highlighting the necessity of low-burden user feedback. To ensure fairness, LPO generates a total number of images equal to the maximum cumulative number produced across all HRFD interaction rounds. 
Notably, we explicitly exclude text-driven methods (e.g., PromptSculptor). While effective for users with clear intents and specialized vocabulary, such methods are fundamentally unsuitable for our problem setting. HRFD specifically targets the ``intention--expression gap,'' where users struggle to verbalize implicit preferences, making a strict low-cognitive-load paradigm essential.

\textbf{Evaluation Metrics.}
We evaluate the methods across three dimensions: efficiency, alignment, and visual quality. 

\textbf{1. Efficiency:}
\textbf{Interaction Rounds (Rounds)} measures convergence efficiency, defined as the number of rounds required for the user to reach satisfaction. Lower values indicate faster intent alignment. Sessions failing to converge are capped at the maximum limit ($T_{max} = 15$).

\textbf{2. Alignment Metrics:}
These metrics quantify the proximity to the user's intent.
\begin{itemize}[noitemsep, topsep=0pt, leftmargin=1em]
    \item \textbf{CLIP Score:} Measures high-level semantic similarity between the generated image and the target ground truth.
    \item \textbf{LPIPS:} Assesses perceptual distance using deep feature embeddings, capturing fine-grained visual and textural alignment.
    \item \textbf{SSIM:} Quantifies low-level pixel alignment and structural fidelity.
\end{itemize}

\textbf{3. Visual Quality:}
\textbf{Aesthetic Score} is evaluated via the pre-trained LAION aesthetic predictor. This verifies that alignment does not degrade the foundational artistic quality.

\subsection{Performance Evaluation}
\label{sec:performance evaluation}
To comprehensively evaluate the HRFD's performance and robustness, we design two distinct experimental scenarios: Objective Preference Alignment (with an explicit visual anchor) and Subjective Preference Discovery (based on users' latent mental intent). The experimental evaluation was conducted by 15 human volunteers.

\textbf{Scenario I: Objective Preference Alignment.}
In this scenario, participants were provided with a feature-rich target image as the Ground Truth (GT) alongside a concise base prompt containing only the visual subject. To ensure the GT images resided entirely within the reachable generative manifold of the underlying model, we constructed them via a two-stage pipeline: a foundation model first generated feature-rich prompts (meta-prompt in Appendix \ref{app:gt_generation_metaprompt}). These detailed prompts were subsequently rendered by the FLUX model to establish the GT references. We deliberately employed the same base model (FLUX) for both GT generation and the iterative alignment process. This homogeneous design is essential to rigorously isolate our framework's pure search efficiency; it ensures that any failure to reconstruct the GT stems from the alignment algorithm itself, rather than a cross-model setup's intrinsic inability to render foreign latent distributions. Furthermore, the GT prompts were generated in open-ended natural language without restriction to HRFD's predefined vocabulary, ensuring our framework holds no unfair advantage over APPO in approximating the GT.

Each participant evaluated 10 distinct cases. During evaluation, the GT image was consistently presented as the target. For the iterative algorithms (HRFD and APPO), participants guided the generation using multi-round, low-burden feedback to align with the GT. In contrast, for the LPO baseline, they simply selected the best-matching candidate from the initial generated pool.
\begin{table}[!htbp]
    \caption{Quantitative comparison of HRFD against baselines. Results are reported as Mean ($\pm$SD). The best performance is in \textbf{bold}. The metrics are grouped by their primary purpose. Arrows indicate preferred direction ($\uparrow$ for higher is better, $\downarrow$ for lower is better). The symbols $^{*}$ and $^{**}$ indicate statistically significant improvements over the strongest baseline (APPO) at $p < 0.05$ and $p < 0.01$, respectively.}
    \label{tab:quantitative_comparison}
    \centering
    % 【重点修改】将 arraystretch 从 1.2 提升到了 1.4，大幅增加行间距，让表格更透气
    \renewcommand{\arraystretch}{1.0}
    \begin{tabular}{@{}llccc@{}}
        \toprule
        \textbf{Metric Group} & \textbf{Metric} & \textbf{LPO} & \textbf{APPO} & \textbf{HRFD (Ours)} \\
        \midrule
        
        \textbf{Efficiency} & Rounds $\downarrow$ & 15.00 $\pm$ 0.0000 & 13.93 $\pm$ 1.7348 & \textbf{11.20 $\pm$ 1.4230$^{**}$} \\
        \midrule
        
        \multirow{3}{*}{\textbf{Alignment}} 
        & CLIP $\uparrow$    & 0.7909 $\pm$ 0.0828 & 0.8309 $\pm$ 0.0744 & \textbf{0.8649 $\pm$ 0.0447$^{**}$} \\
        & LPIPS $\downarrow$ & 0.7623 $\pm$ 0.0680 & 0.7285 $\pm$ 0.0685 & \textbf{0.6890 $\pm$ 0.0573$^{**}$} \\
        & SSIM $\uparrow$    & 0.1851 $\pm$ 0.1936 & 0.2160 $\pm$ 0.1766 & \textbf{0.3137 $\pm$ 0.1583$^{**}$} \\
        \midrule
        
        \textbf{Visual Quality} & Aesthetic $\uparrow$ & 6.7805 $\pm$ 0.0381 & 6.8185 $\pm$ 0.0398 & \textbf{6.9249 $\pm$ 0.0379}\\
        
        \bottomrule
    \end{tabular}
\end{table}

\begin{figure}[!htbp]
  \centering
  \begin{minipage}{0.32\linewidth}
    \centering
    \includegraphics[width=\linewidth]{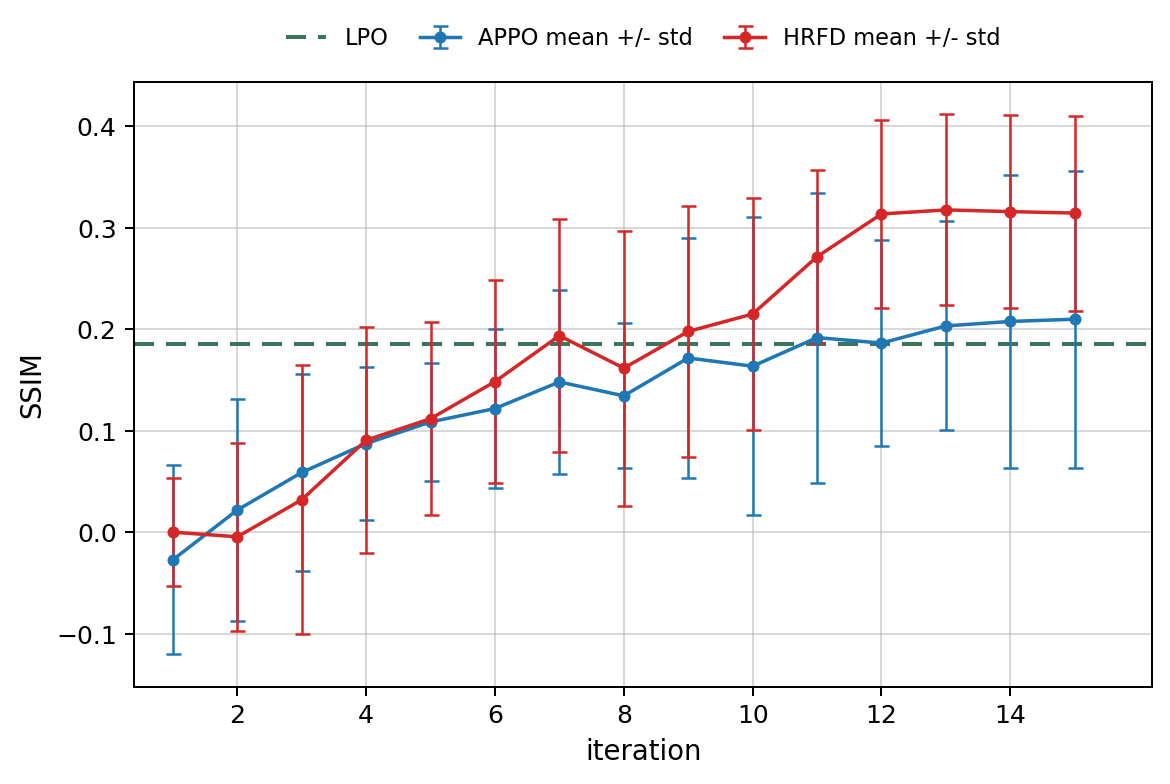}
    
    \vspace{3pt}
    \small (a) SSIM
  \end{minipage}
  \hfill
  \begin{minipage}{0.32\linewidth}
    \centering
    \includegraphics[width=\linewidth]{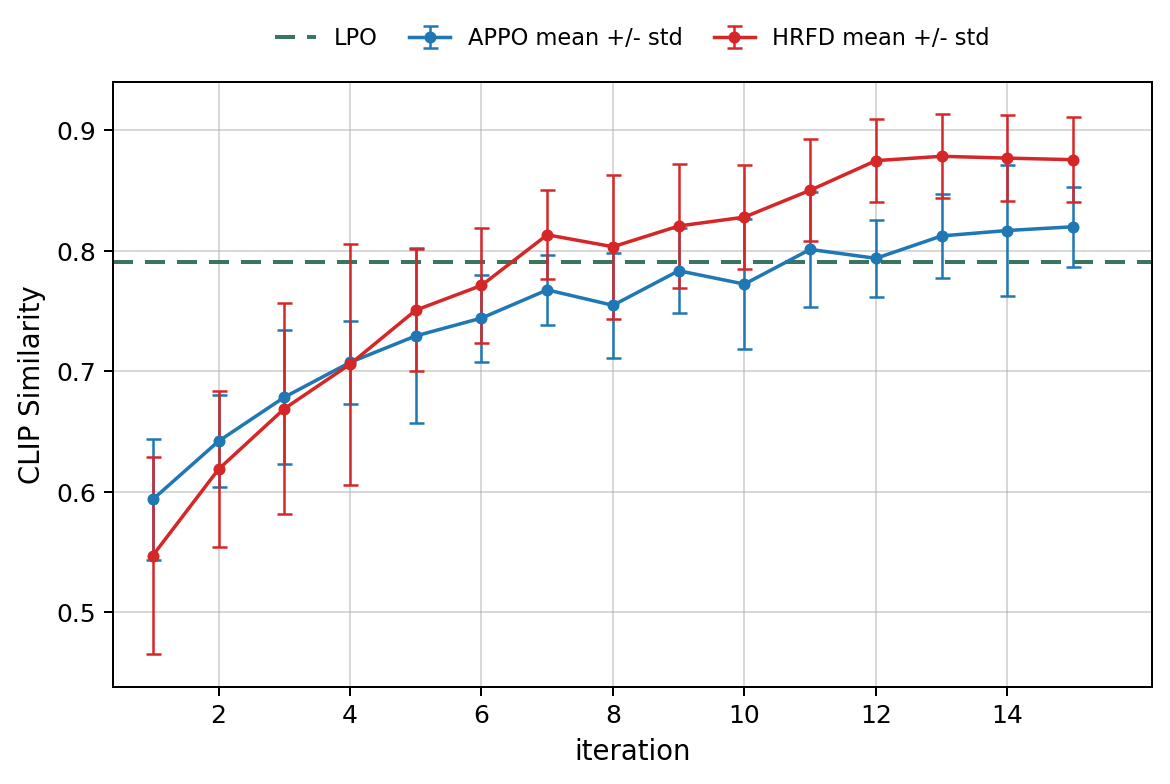}
    
    \vspace{3pt}
    \small (b) CLIP Similarity
  \end{minipage}
  \hfill
  \begin{minipage}{0.32\linewidth}
    \centering
    \includegraphics[width=\linewidth]{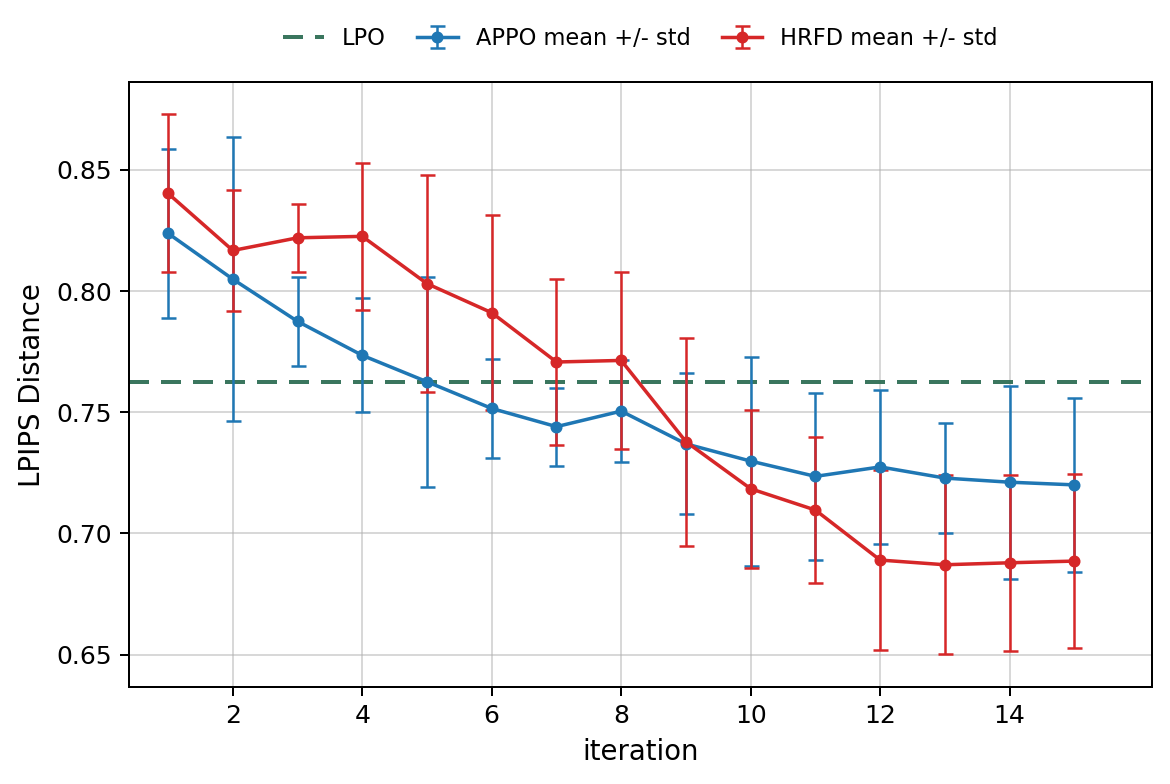}
    
    \vspace{3pt}
    \small (c) LPIPS Distance
  \end{minipage}
  \caption{Quantitative performance evolution across interaction rounds in Scenario I. The curves report the mean and standard deviation for SSIM, CLIP Similarity, and LPIPS Distance, demonstrating the more stable and superior convergence of HRFD.}
  \label{fig:iterative_curves}
\end{figure}

Table~\ref{tab:quantitative_comparison} and Figure~\ref{fig:iterative_curves} present the quantitative results and convergence evolution ($p < 0.01$). First, we evaluate convergence efficiency. HRFD reaches user satisfaction in 11.20 rounds on average, compared to 13.93 for APPO. The LPO baseline, which lacks user feedback, defaults to exhausting the full interaction budget (15.00) without converging. This empirically confirms our core premise: lightweight user feedback is strictly necessary to resolve ambiguous visual intent, which pure language model rewriting cannot achieve alone.
Second, regarding alignment performance, the advantage of HRFD over APPO grows monotonically with metric granularity: $+4.1\%$ at the semantic level (CLIP), $+5.4\%$ at the perceptual level (LPIPS), and $+45.2\%$ at the structural level (SSIM). Although absolute SSIM is naturally low in unconstrained image generation due to spatial variance, the significant relative improvement over baselines confirms HRFD's effectiveness in structural alignment. During experiments, participants consistently reported selecting one candidate for its composition and another for its lighting to fuse these features. However, APPO fails this integration. From APPO's perspective, these selections often introduce contradictory signals (e.g., opposing compositional structures), obstructing its FM's inference of true preferences. In contrast, HRFD's robust statistical formulations decouple each feature independently, quantifying the distribution divergence between liked and disliked sets to isolate true preferences from such conflicting signals, enabling simultaneous alignment across all dimensions.
Finally, in terms of visual quality, comparable aesthetic scores across all methods verify that HRFD achieves intent alignment without degrading the underlying model’s foundational generative quality. Extended qualitative results are provided in Appendix~\ref{app:Qualitative Analysis}.

\textbf{Scenario II: Subjective Preference Discovery.}
Unlike Scenario I, participants annotated candidates based on subjective aesthetic preferences without external references, simulating real-world conditions where users hold clear but not yet concretized visual intents. Each participant completed 10 distinct interaction sessions. Since no ground truth existed, evaluation relied on Interaction Rounds for convergence efficiency, the NASA Task Load Index (NASA-TLX)~\cite{NASA-TLX} for cognitive load, and the Creativity Support Index (CSI)~\cite{CSI} for expressive effectiveness.

\begin{figure}[!htbp]
  \centering
  \includegraphics[width=\linewidth]{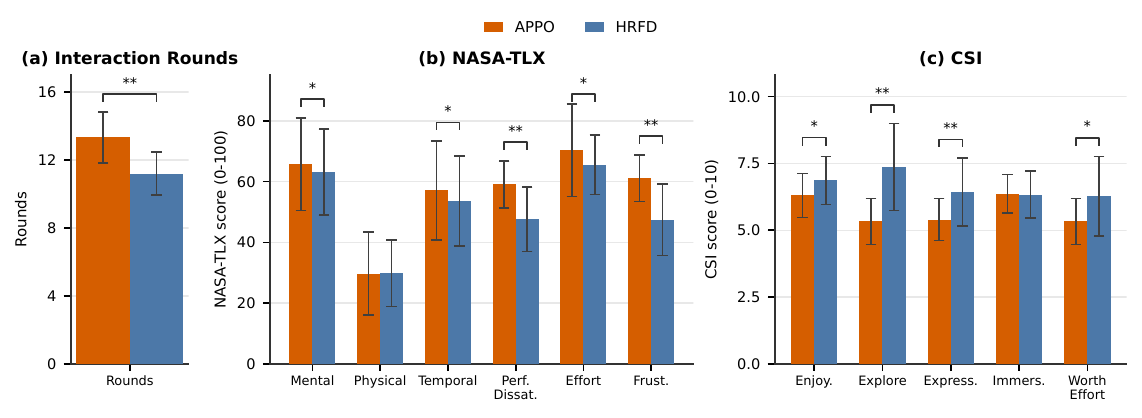} 
  \vspace{-15pt} % <<--- 添加这一行，通过调整这个值（如 -5pt 到 -15pt）来精确控制空白大小
\caption{Subjective evaluation results for Scenario II. HRFD demonstrates superior efficiency (Rounds), lower cognitive load (NASA-TLX), and higher creativity support (CSI) compared to the baseline (*$p < 0.05$, **$p < 0.01$).}  \label{fig:scenario2_results}
\end{figure}

As illustrated in Figure~\ref{fig:scenario2_results}, HRFD achieves statistically significant improvements over APPO across all primary metrics ($p < 0.05$, Wilcoxon signed-rank test; detailed statistics in Appendix~\ref{app:statistical_results}). As expected, Physical Demand ($p = 0.9320$) and Immersion ($p = 0.7907$) show no significant difference, as both methods share an identical click-based interaction paradigm. The significant improvements across all other primary metrics successfully validate the overall effectiveness of HRFD.

\subsection{Ablation Study}
\label{sec:study}
To verify the necessity of each core component within HRFD, we conducted an ablation study comprising three variants. \textbf{1) w/o Hierarchy:} We flattened the three-tier feature architecture, forcing the system to evaluate and update all variables simultaneously during each interaction round, thereby removing the coarse-to-fine locking mechanism. \textbf{2) w/o History:} We disabled the global preference memory for both discrete and ordinal features, restricting preference inference exclusively to isolated, single-round feedback and preventing any cross-round information accumulation. \textbf{3) w/o Weighting:} We eliminated the dynamic weighting mechanism, assigning uniform weights to all interaction rounds regardless of their actual information quality.

\begin{table}[!htbp]
    \caption{Ablation study results. The best performance is in \textbf{bold}. 
    Arrows indicate preferred direction ($\uparrow$ higher is better, 
    $\downarrow$ lower is better).}
    \label{tab:ablation}
    \centering
    \renewcommand{\arraystretch}{1.0}
    \small
    \begin{tabular}{@{}lccccc@{}}
        \toprule
        \textbf{Variant} & \textbf{Rounds $\downarrow$} & \textbf{CLIP $\uparrow$} & \textbf{LPIPS $\downarrow$} & \textbf{SSIM $\uparrow$} & \textbf{Aesthetic $\uparrow$} \\
        \midrule
        w/o Hierarchy & 14.63 & 0.7553 & 0.7932 & 0.1491 & \textbf{7.0547} \\
        w/o History   & 14.20 & 0.8196 & 0.7610 & 0.1903& 6.9007 \\
        w/o Weighting & 12.93 & 0.8334 & 0.7346 & 0.2186& 6.9500 \\
        \midrule
        \textbf{HRFD (Ours)} & \textbf{11.20} & \textbf{0.8649} & \textbf{0.6890} & \textbf{0.3137}& 6.9249 \\
        \bottomrule
    \end{tabular}
\end{table}

All variants were evaluated following the identical human-in-the-loop experimental protocol used in Scenario I. As summarized in Table~\ref{tab:ablation}, HRFD consistently outperforms all its ablated variants across both convergence efficiency and alignment accuracy, validating the structural necessity of each proposed component. Detailed experimental analysis is provided in Appendix~\ref{app:ablation_analysis}.

\section{Conclusion and Limitations}
\label{sec:conclusion}
% This paper introduces the HRFD framework to bridge the intention-expression gap in multi-dimensional feature alignment. Existing methods rely on holistic Foundation Model inference, which suffers from severe inference overload and fails to resolve conflicting user signals. HRFD overcomes these bottlenecks by structuring preferences into a three-tier hierarchy for progressive, coarse-to-fine convergence. By decomposing complex reasoning into independent single-feature tasks and quantifying preferences via the distribution divergence between liked and disliked sets, HRFD achieves a robust, transparent alignment that FMs inherently lack. Extensive evaluations confirm HRFD's significant superiority in both interaction efficiency and alignment accuracy.
% Despite its effectiveness, HRFD's reliance on a predefined, static feature repository introduces an Out-of-Vocabulary (OOV) bottleneck. While this closed-vocabulary approach enables rigorous mathematical tracking and prevents FM hallucinations, it cannot dynamically capture highly niche attributes, effectively transforming the intention-expression gap into a dictionary coverage gap. Future work will explore a Dynamic Feature Repository that autonomously discovers and integrates emergent features based on continuous user feedback. By mathematically tracking these emergent attributes, this adaptation aims to resolve the coverage gap, fully unleashing the open-world generative potential of diffusion models.

This paper introduces HRFD to bridge the intention-expression gap in multi-dimensional alignment. Unlike holistic Foundation Model inference methods that suffer from inference overload and fail under conflicting signals, HRFD structures preferences into a three-tier hierarchy for coarse-to-fine convergence. By decomposing reasoning into single-feature tasks and statistically quantifying preference divergence, HRFD achieves a robust, transparent alignment that FMs inherently lack. Extensive evaluations confirm its superior efficiency and accuracy for universal visual features. However, HRFD’s reliance on a static feature repository introduces an Out-of-Vocabulary bottleneck. This constraint prevents the capture of highly niche attributes, effectively transforming the intention-expression gap into a dictionary coverage gap. Future work will explore a Dynamic Feature Repository driven by FMs. By leveraging extensive prior knowledge to analyze historical feedback, FMs can adaptively update this repository in real-time: pruning unconcerned dimensions and replacing them with newly discovered features of interest, while simultaneously refining the specific values of confirmed preference directions for highly precise alignment. This dynamic framework will fully unleash the open-world generative potential of diffusion models.

\FloatBarrier
\bibliographystyle{plainnat}
\bibliography{references}

@inproceedings{rombach2022high,
  title={High-resolution image synthesis with latent diffusion models},
  author={Rombach, Robin and Blattmann, Andreas and Lorenz, Dominik and Esser, Patrick and Ommer, Bj{\"o}rn},
  booktitle={Proceedings of the IEEE/CVF Conference on Computer Vision and Pattern Recognition (CVPR)},
  pages={10684--10695},
  year={2022}
}

@misc{dai2023emu,
      title={Emu: Enhancing Image Generation Models Using Photogenic Needles in a Haystack}, 
      author={Xiaoliang Dai and Ji Hou and Chih-Yao Ma and Sam Tsai and Jialiang Wang and Rui Wang and Peizhao Zhang and Simon Vandenhende and Xiaofang Wang and Abhimanyu Dubey and Matthew Yu and Abhishek Kadian and Filip Radenovic and Dhruv Mahajan and Kunpeng Li and Yue Zhao and Vladan Petrovic and Mitesh Kumar Singh and Simran Motwani and Yi Wen and Yiwen Song and Roshan Sumbaly and Vignesh Ramanathan and Zijian He and Peter Vajda and Devi Parikh},
      year={2023},
      eprint={2309.15807},
      archivePrefix={arXiv},

}

@misc{podell2023sdxl,
      title={SDXL: Improving Latent Diffusion Models for High-Resolution Image Synthesis}, 
      author={Dustin Podell and Zion English and Kyle Lacey and Andreas Blattmann and Tim Dockhorn and Jonas Müller and Joe Penna and Robin Rombach},
      year={2023},
      eprint={2307.01952},
      archivePrefix={arXiv},

}

@article{feng2022training,
  title={Training-free structured diffusion guidance for compositional text-to-image synthesis},
  author={Feng, Weixi and He, Xuehai and Fu, Tsu-Jui and Jampani, Varun and Akula, Arjun and Narayana, Pradyumna and Basu, Sugato and Wang, Xin Eric and Wang, William Yang},
  journal={arXiv preprint arXiv:2212.05032},
  year={2022}
}

@article{hertz2022prompt,
  title={Prompt-to-prompt image editing with cross attention control},
  author={Hertz, Amir and Mokady, Ron and Tenenbaum, Jay and Aberman, Kfir and Pritch, Yael and Cohen-Or, Daniel},
  journal={arXiv preprint arXiv:2208.01626},
  year={2022}
}

@article{chefer2023attend,
  title={Attend-and-excite: Attention-based semantic guidance for text-to-image diffusion models},
  author={Chefer, Hila and Alaluf, Yuval and Vinker, Yael and Wolf, Lior and Cohen-Or, Daniel},
  journal={ACM Transactions on Graphics (TOG)},
  volume={42},
  number={4},
  pages={1--10},
  year={2023},
  publisher={ACM New York, NY, USA}
}

@inproceedings{xu2023imagereward,
  title={{ImageReward}: Learning and evaluating human preferences for text-to-image generation},
  author={Xu, Jiazheng and Liu, Xiao and Wu, Yuchen and Tong, Yuxuan and Li, Qinkai and Ding, Ming and Tang, Jie and Dong, Yuxiao},
  booktitle={Advances in Neural Information Processing Systems (NeurIPS)},
  volume={36},
  pages={15903--15935},
  year={2023}
}

@inproceedings{an2025agfsync,
  title={{Agfsync}: Leveraging {AI}-generated feedback for preference optimization in text-to-image generation},
  author={An, Jingkun and Zhu, Yinghao and Li, Zongjian and Zhou, Enshen and Feng, Haoran and Huang, Xijie and Chen, Bohua and Shi, Yemin and Pan, Chengwei},
  booktitle={Proceedings of the AAAI Conference on Artificial Intelligence},
  volume={39},
  number={2},
  pages={1746--1754},
  year={2025}
}

@article{datta2008image,
  title={Image retrieval: Ideas, influences, and trends of the new age},
  author={Datta, Ritendra and Joshi, Dhiraj and Li, Jia and Wang, James Z},
  journal={ACM Computing Surveys (CSUR)},
  volume={40},
  number={2},
  pages={1--60},
  year={2008},
  publisher={ACM New York, NY, USA}
}

@techreport{betker2023improving,
  title={Improving image generation with better captions},
  author={Betker, James and Goh, Gabriel and Jing, Li and Brooks, Tim and Wang, Jianfeng and Li, Linjie and Ouyang, Long and Zhuang, Juntang and Lee, Joyce and Guo, Yufei and others},
  institution={OpenAI},
  year={2023},
  url={https://cdn.openai.com/papers/dall-e-3.pdf}
}

@inproceedings{sun2025dreamsync,
  title={{DreamSync}: Aligning text-to-image generation with image understanding feedback},
  author={Sun, Jiao and Fu, Deqing and Hu, Yushi and Wang, Su and Rassin, Royi and Juan, Da-Cheng and Alon, Dana and Herrmann, Charles and Van Steenkiste, Sjoerd and Krishna, Ranjay and Rashtchian, Cyrus},
  booktitle={Proceedings of the Conference of the Nations of the Americas Chapter of the Association for Computational Linguistics (NAACL)},
  pages={5920--5945},
  year={2025}
}

@article{broilo2010stochastic,
  author={Broilo, Mattia and De Natale, Francesco G. B.},
  journal={IEEE Transactions on Multimedia}, 
  title={A Stochastic Approach to Image Retrieval Using Relevance Feedback and Particle Swarm Optimization}, 
  year={2010},
  volume={12},
  number={4},
  pages={267-277},

}

@inproceedings{hao2022optimizing,
  title={Optimizing prompts for text-to-image generation},
  author={Hao, Yaru and Chi, Zewen and Dong, Li and Wei, Furu},
  booktitle={Advances in Neural Information Processing Systems (NeurIPS)},
  volume={35},
  pages={17182--17194},
  year={2022}
}

@inproceedings{hong2000incorporate,
  title={Incorporate support vector machines to content-based image retrieval with relevant feedback},
  author={Hong, P. and Tian, Q. and Huang, T. S.},
  booktitle={Proceedings of the IEEE International Conference on Image Processing (ICIP)},
  volume={3},
  pages={750--753},
  year={2000},
  organization={IEEE}
}

@inproceedings{sohl2015deep,
  title={Deep unsupervised learning using nonequilibrium thermodynamics},
  author={Sohl-Dickstein, Jascha and Weiss, Eric A and Maheswaranathan, Niru and Ganguli, Surya},
  booktitle={Proceedings of the 32nd International Conference on Machine Learning (ICML)},
  pages={2256--2265},
  volume = 	 {37},
  year={2015}
}

@article{kanimozhi2015integrated,
  title={An integrated approach to region based image retrieval using firefly algorithm and support vector machine},
  author={Kanimozhi, T. and Latha, K.},
  journal={Neurocomputing},
  volume={151},
  pages={1099--1111},
  year={2015},
  publisher={Elsevier}
}

@inproceedings{mackie2023generative,
  title={Generative relevance feedback with large language models},
  author={Mackie, Iain and Chatterjee, Shubham and Dalton, Jeffrey},
  booktitle={Proceedings of the 46th International ACM SIGIR Conference on Research and Development in Information Retrieval},
year = {2023},
pages = {2026–2031},
numpages = {6}
}

@article{mahmood2022hybrid,
  title={Hybrid evolutionary algorithm based relevance feedback approach for image retrieval},
  author={Mahmood, A. and Imran, M. and Irtaza, A. and Abbas, Q. and Dhahri, H.},
  journal={Computers, Materials \& Continua},
  volume={70},
  number={1},
  pages={963--979},
  year={2022}
}

@incollection{rocchio1971relevance,
  title={Relevance feedback in information retrieval},
  author={Rocchio, Joseph J},
  booktitle={The SMART Retrieval System: Experiments in Automatic Document Processing},
  editor={Salton, G.},
  pages={313--323},
  year={1971},
  publisher={Prentice-Hall}
}

@article{rui1998relevance,
  title={Relevance feedback: A power tool for interactive content-based image retrieval},
  author={Rui, Yong and Huang, Thomas S and Mehrotra, Sharad and Ortega, Michael},
  journal={IEEE Transactions on Circuits and Systems for Video Technology},
  volume={8},
  number={5},
  pages={644--655},
  year={1998},
  publisher={IEEE}
}

@inproceedings{sharma2025can,
  title={Can relevance feedback, conversational search and foundation models work together for interactive video search and exploration?},
  author={Sharma, U. and Rudinac, S. and Khan, O. S. and J{\'o}nsson, B. {\TH}.},
  booktitle={Proceedings of the IEEE/CVF Conference on Computer Vision and Pattern Recognition Workshops (CVPRW)},
    year      = {2025},
    pages     = {3779-3788}
}

@article{tao2006asymmetric,
  title={Asymmetric bagging and random subspace for support vector machines-based relevance feedback in image retrieval},
  author={Tao, Dacheng and Tang, Xiaoou and Li, Xuelong and Wu, Xindong},
  journal={IEEE Transactions on Pattern Analysis and Machine Intelligence (TPAMI)},
  volume={28},
  number={7},
  pages={1088--1099},
  year={2006},
  publisher={IEEE}
}

@article{tong2001support,
  title={Support vector machine active learning with applications to text classification},
  author={Tong, Simon and Koller, Daphne},
  journal={Journal of Machine Learning Research},
  volume={2},
  number={Nov},
  pages={45--66},
  year={2001}
}

@inproceedings{zhang2005using,
  title={Using high dimensional indexes to support relevance feedback based interactive images retrieval},
  author={Zhang, J. and Zhou, X. and Wang, W. and Shi, B. and Pei, J.},
  booktitle={Proceedings of the 31st International Conference on Very Large Data Bases (VLDB)},
  pages={1211--1214},
  year={2005}
}

@inproceedings{zou2019reinforcement,
  title={Reinforcement learning to optimize long-term user engagement in recommender systems},
  author={Zou, L. and Xia, L. and Ding, Z. and Song, J. and Liu, W. and Yin, D.},
  booktitle={Proceedings of the 25th ACM SIGKDD International Conference on Knowledge Discovery \& Data Mining (KDD)},
  pages={2810--2818},
  year={2019},
  organization={ACM}
}

@article{zhou2003relevance,
  title={Relevance feedback in image retrieval: A comprehensive review},
  author={Zhou, Xiang Sean and Huang, Thomas S},
  journal={Multimedia Systems},
  volume={8},
  number={6},
  pages={536--544},
  year={2003},
  publisher={Springer}
}

@inproceedings{kong2016photo,
  author    = {Kong, Shu and Shen, Xiaohui and Lin, Zhe and Mech, Radomir and Fowlkes, Charless},
  title     = {Photo aesthetics ranking network with attributes and content adaptation},
  booktitle = {Proceedings of the European Conference on Computer Vision (ECCV)},
  year      = {2016},
  volume ={9905},
  publisher = {Springer},
  pages     = {662--679}
}

@misc{segalis2023picture,
      title={A Picture is Worth a Thousand Words: Principled Recaptioning Improves Image Generation}, 
      author={Eyal Segalis and Dani Valevski and Danny Lumen and Yossi Matias and Yaniv Leviathan},
      year={2023},
      eprint={2310.16656},
      archivePrefix={arXiv},
}

@inproceedings{DDPO,
 author = {Black, Kevin and Janner, Michael and Du, Yilun and Kostrikov, Ilya and Levine, Sergey},
 booktitle = {International Conference on Learning Representations},
 editor = {B. Kim and Y. Yue and S. Chaudhuri and K. Fragkiadaki and M. Khan and Y. Sun},
 pages = {4965--4987},
 title = {Training Diffusion Models with Reinforcement Learning},
 volume = {2024},
 year = {2024}
}

@InProceedings{Diffusion-DPO,
    author    = {Wallace, Bram and Dang, Meihua and Rafailov, Rafael and Zhou, Linqi and Lou, Aaron and Purushwalkam, Senthil and Ermon, Stefano and Xiong, Caiming and Joty, Shafiq and Naik, Nikhil},
    title     = {Diffusion Model Alignment Using Direct Preference Optimization},
    booktitle = {Proceedings of the IEEE/CVF Conference on Computer Vision and Pattern Recognition (CVPR)},
    month     = {June},
    year      = {2024},
    pages     = {8228-8238}
}

@inproceedings{
IterComp,
title={IterComp: Iterative Composition-Aware Feedback Learning from Model Gallery for Text-to-Image Generation},
author={Xinchen Zhang and Ling Yang and Guohao Li and YaQi Cai and xie jiake and Yong Tang and Yujiu Yang and Mengdi Wang and Bin CUI},
booktitle={The Thirteenth International Conference on Learning Representations},
year={2025},
url={https://openreview.net/forum?id=4w99NAikOE}
}

@misc{Twin-Co,
      title={Twin Co-Adaptive Dialogue for Progressive Image Generation}, 
      author={Jianhui Wang and Yangfan He and Yan Zhong and Xinyuan Song and Jiayi Su and Yuheng Feng and Ruoyu Wang and Hongyang He and Wenyu Zhu and Xinhang Yuan and Miao Zhang and Keqin Li and Jiaqi Chen and Tianyu Shi and Xueqian Wang},
      year={2026},
      eprint={2504.14868},
      archivePrefix={arXiv},
}

@inproceedings{
Pasta,
title={Preference Adaptive and Sequential Text-to-Image Generation},
author={Ofir Nabati and Guy Tennenholtz and ChihWei Hsu and Moonkyung Ryu and Deepak Ramachandran and Yinlam Chow and Xiang Li and Craig Boutilier},
booktitle={Forty-second International Conference on Machine Learning},
year={2025},
url={https://openreview.net/forum?id=LCr6CIAEye}
}

@inproceedings{
DNO,
title={Inference-Time Alignment of Diffusion Models with Direct Noise Optimization},
author={Zhiwei Tang and Jiangweizhi Peng and Jiasheng Tang and Mingyi Hong and Fan Wang and Tsung-Hui Chang},
booktitle={Forty-second International Conference on Machine Learning},
year={2025},
url={https://openreview.net/forum?id=JpbqiD7n9r}
}

@inproceedings{
DAS,
title={Test-time Alignment of Diffusion Models without Reward Over-optimization},
author={Sunwoo Kim and Minkyu Kim and Dongmin Park},
booktitle={The Thirteenth International Conference on Learning Representations},
year={2025},
url={https://openreview.net/forum?id=vi3DjUhFVm}
}

@inproceedings{
Demon,
title={Training-Free Diffusion Model Alignment with Sampling Demons},
author={Po-Hung Yeh and Kuang-Huei Lee and Jun-cheng Chen},
booktitle={The Thirteenth International Conference on Learning Representations},
year={2025},
url={https://openreview.net/forum?id=tfemquulED}
}

@misc{SIDiffAgent,
      title={SIDiffAgent: Self-Improving Diffusion Agent}, 
      author={Shivank Garg and Ayush Singh and Gaurav Kumar Nayak},
      year={2026},
      eprint={2602.02051},
      archivePrefix={arXiv},
}

@misc{RAVEL,
      title={RAVEL: Rare Concept Generation and Editing via Graph-driven Relational Guidance}, 
      author={Kavana Venkatesh and Yusuf Dalva and Ismini Lourentzou and Pinar Yanardag},
      year={2025},
      eprint={2412.09614},
      archivePrefix={arXiv},
}

@misc{RAISE,
      title={RAISE: Requirement-Adaptive Evolutionary Refinement for Training-Free Text-to-Image Alignment}, 
      author={Liyao Jiang and Ruichen Chen and Chao Gao and Di Niu},
      eprint={2603.00483},
      archivePrefix={arXiv},
            year={2026},
}

@inproceedings{APPO,
author = {Li, Zhipeng and Liao, Yi-Chi and Holz, Christian},
title = {Preference-Guided Prompt Optimization for Text-to-Image Generation},
year = {2026},
isbn = {9798400722783},
publisher = {Association for Computing Machinery},
address = {New York, NY, USA},
booktitle = {Proceedings of the 2026 CHI Conference on Human Factors in Computing Systems},
location = {
},
series = {CHI '26}
}

@inproceedings{PromptCharm, series={CHI ’24},
   title={PromptCharm: Text-to-Image Generation through Multi-modal Prompting and Refinement},
   booktitle={Proceedings of the CHI Conference on Human Factors in Computing Systems},
   publisher={ACM},
   author={Wang, Zhijie and Huang, Yuheng and Song, Da and Ma, Lei and Zhang, Tianyi},
   year={2024},
   month=may, pages={1–21},
   collection={CHI ’24} }

@misc{DSPy,
      title={DSPy: Compiling Declarative Language Model Calls into Self-Improving Pipelines}, 
      author={Omar Khattab and Arnav Singhvi and Paridhi Maheshwari and Zhiyuan Zhang and Keshav Santhanam and Sri Vardhamanan and Saiful Haq and Ashutosh Sharma and Thomas T. Joshi and Hanna Moazam and Heather Miller and Matei Zaharia and Christopher Potts},
      year={2023},
      eprint={2310.03714},
      archivePrefix={arXiv},
}

@misc{Promptimizer,
      title={Promptimizer: User-Led Prompt Optimization for Personal Content Classification}, 
      author={Leijie Wang and Kathryn Yurechko and Amy X. Zhang},
      year={2025},
      eprint={2510.09009},
      archivePrefix={arXiv},
}

@misc{dalle2,
      title={Hierarchical Text-Conditional Image Generation with CLIP Latents}, 
      author={Aditya Ramesh and Prafulla Dhariwal and Alex Nichol and Casey Chu and Mark Chen},
      year={2022},
      eprint={2204.06125},
      archivePrefix={arXiv},
}

@ARTICLE{PrompTHis,
  author={Guo, Yuhan and Shao, Hanning and Liu, Can and Xu, Kai and Yuan, Xiaoru},
  journal={IEEE Transactions on Visualization and Computer Graphics}, 
  title={PrompTHis: Visualizing the Process and Influence of Prompt Editing During Text-to-Image Creation}, 
  year={2025},
  volume={31},
  number={9},
  pages={4547-4559},
  }

@inproceedings{ConstitutionMaker,
author = {Petridis, Savvas and Wedin, Benjamin D and Wexler, James and Pushkarna, Mahima and Donsbach, Aaron and Goyal, Nitesh and Cai, Carrie J and Terry, Michael},
title = {ConstitutionMaker: Interactively Critiquing Large Language Models by Converting Feedback into Principles},
year = {2024},
isbn = {9798400705083},
publisher = {Association for Computing Machinery},
address = {New York, NY, USA},
booktitle = {Proceedings of the 29th International Conference on Intelligent User Interfaces},
pages = {853–868},
numpages = {16},
location = {Greenville, SC, USA},
series = {IUI '24}
}

@book{cohen2013statistical,
  title={Statistical power analysis for the behavioral sciences},
  author={Cohen, Jacob},
  year={2013},
  publisher={routledge}
}

@article{navon1977,
title = {Forest before trees: The precedence of global features in visual perception},
journal = {Cognitive Psychology},
volume = {9},
number = {3},
pages = {353-383},
year = {1977},
author = {David Navon},
}

@article{hegde2008,
title = {Time course of visual perception: Coarse-to-fine processing and beyond},
journal = {Progress in Neurobiology},
volume = {84},
number = {4},
pages = {405-439},
year = {2008},
author = {Jay Hegdé},
}

@article{dalkey1963,
  author    = {Dalkey, Norman and Helmer, Olaf},
  title     = {An experimental application of the {Delphi} method 
               to the use of experts},
  journal   = {Management Science},
  volume    = {9},
  number    = {3},
  pages     = {458--467},
  year      = {1963}
}

@incollection{NASA-TLX,
title = {Development of NASA-TLX (Task Load Index): Results of Empirical and Theoretical Research},
editor = {Peter A. Hancock and Najmedin Meshkati},
series = {Advances in Psychology},
publisher = {North-Holland},
volume = {52},
pages = {139-183},
year = {1988},
booktitle = {Human Mental Workload},
issn = {0166-4115},
author = {Sandra G. Hart and Lowell E. Staveland},
}

@article{CSI,
author = {Cherry, Erin and Latulipe, Celine},
title = {Quantifying the Creativity Support of Digital Tools through the Creativity Support Index},
year = {2014},
issue_date = {August 2014},
publisher = {Association for Computing Machinery},
address = {New York, NY, USA},
volume = {21},
number = {4},
issn = {1073-0516},
journal = {ACM Trans. Comput.-Hum. Interact.},
month = jun,
articleno = {21},
}

@article{liu2024lost,
  title={Lost in the middle: How language models use long contexts},
  author={Liu, Nelson F and Lin, Kevin and Hewitt, John and Paranjape, Ashwin and Bevilacqua, Michele and Petroni, Fabio and Liang, Percy},
  journal={Transactions of the Association for Computational Linguistics},
  volume={12},
  pages={157--173},
  year={2024},
  publisher={MIT Press}
}

@inproceedings{dziri2023faith,
  title={Faith and fate: Limits of transformers on compositionality},
  author={Dziri, Nouha and Lu, Ximing and Sclar, Melanie and Li, Xiang Lorraine and Jian, Liwei and Lin, Bill Yuchen and West, Peter and Bhagavatula, Chandra and Bras, Ronan Le and Hwang, Jena D and others},
  booktitle={Advances in Neural Information Processing Systems},
  volume={36},
  pages={55152--55198},
  year={2023}
}

@inproceedings{ouyang2022training,
  title={Training language models to follow instructions with human feedback},
  author={Ouyang, Long and Wu, Jeffrey and Jiang, Xu and Almeida, Diogo and Wainwright, Carroll and Mishkin, Pamela and Zhang, Chong and Agarwal, Sandhini and Slama, Katarina and Ray, Alex and others},
  booktitle={Advances in Neural Information Processing Systems},
  volume={35},
  pages={27730--27744},
  year={2022}
}

\appendix

\section{Detailed Feature Standard Repository}
\label{app:appendix_features}

This appendix provides the complete hierarchical feature-standard repository (see Table \ref{tab:table_appendix_features}) utilized in the HRFD framework, extending the summary provided in Section \ref{sec:feature_extraction}. 

\begin{table}[!htbp]
\caption{Complete Hierarchical Feature Standard Repository}
\label{tab:table_appendix_features}
\centering
\footnotesize % 移到附录后，字号可以稍微调大一点，方便审稿人阅读
% 增加全局行高，赋予表格呼吸感
\renewcommand{\arraystretch}{1.3} 
% 优化列间距
\setlength{\tabcolsep}{4pt} 

% 引入 >{\raggedright\arraybackslash} 强制最后一列左对齐，消除异常空格
\begin{tabular}{@{}p{0.12\linewidth}p{0.20\linewidth}p{0.12\linewidth}>{\raggedright\arraybackslash}p{0.50\linewidth}@{}}
\toprule
\textbf{Hierarchy} & \textbf{Feature} & \textbf{Type} & \textbf{Values / Levels} \\
\midrule
\multirow{5}{*}{\makecell[l]{\textbf{L1:}\\\textbf{Macro}}} 
    & Art Style & Discrete & realistic, illustration, flat\_illustration, cartoon, anime, 3d\_render, pixel\_art, watercolor, oil\_painting, sketch, line\_art, minimalist, surreal, collage, mixed\_media \\
    & Style Era & Discrete & classical, vintage, modern, futuristic \\
    & Cultural Style & Discrete & asian, western, african, middle\_eastern, mixed \\
    & Emotion Type & Discrete & serene, joyful, romantic, melancholic, mysterious, tense, playful, ominous, angry\\
    & Environment & Discrete & indoor, outdoor, studio, nature, urban, abstract, fantasy, mixed \\
\midrule
\multirow{12}{*}{\makecell[l]{\textbf{L2:}\\\textbf{Meso}}} 
    & Color Palette & Discrete & monochrome, limited\_palette, analogous, complementary, triadic, varied \\
    & Light Source & Discrete & natural, studio, neon, ambient \\
    & Lighting Design & Discrete & flat, soft\_diffused, backlit, rim, dramatic \\
    & Framing & Ordinal & extreme\_closeup $\rightarrow$ closeup $\rightarrow$ medium $\rightarrow$ wide $\rightarrow$ extreme\_wide \\
    & Perspective & Discrete & flat, eye\_level, low\_angle, high\_angle, birds\_eye, isometric \\
    & Layout & Discrete & centered, rule\_of\_thirds, symmetrical, asymmetrical, diagonal, radial, scattered \\
    & Visual Flow & Discrete & static, dynamic \\
    & Depth Sense & Ordinal & flat $\rightarrow$ shallow $\rightarrow$ moderate $\rightarrow$ deep $\rightarrow$ extreme\_deep \\
    & Negative Space & Ordinal & cramped $\rightarrow$ minimal $\rightarrow$ balanced $\rightarrow$ abundant $\rightarrow$ dominant \\
    & Background Mode & Discrete & solid, gradient, patterned, abstract\_graphic, scenic \\
    & Scene Complexity & Ordinal & minimal $\rightarrow$ simple $\rightarrow$ moderate $\rightarrow$ detailed $\rightarrow$ complex \\
    & Emotion Intensity & Ordinal & subtle $\rightarrow$ mild $\rightarrow$ moderate $\rightarrow$ strong $\rightarrow$ overwhelming \\
\midrule
\multirow{8}{*}{\makecell[l]{\textbf{L3:}\\\textbf{Micro}}} 
    & Temperature & Ordinal & cold $\rightarrow$ cool $\rightarrow$ neutral $\rightarrow$ warm $\rightarrow$ hot \\
    & Saturation & Ordinal & desaturated $\rightarrow$ muted $\rightarrow$ moderate $\rightarrow$ vibrant $\rightarrow$ oversaturated \\
    & Brightness & Ordinal & dark $\rightarrow$ dim $\rightarrow$ medium $\rightarrow$ bright $\rightarrow$ very\_bright \\
    & Contrast & Ordinal & low $\rightarrow$ medium\_low $\rightarrow$ medium $\rightarrow$ high $\rightarrow$ extreme \\
    & Background Blur & Ordinal & sharp $\rightarrow$ slightly\_blurred $\rightarrow$ heavily\_blurred \\
    & Detail Level & Ordinal & low $\rightarrow$ medium\_low $\rightarrow$ medium $\rightarrow$ high $\rightarrow$ ultra \\
    & Texture Rendering & Ordinal & flat $\rightarrow$ subtle $\rightarrow$ detailed $\rightarrow$ hyper\_realistic \\
    & Finish Quality & Ordinal & rough $\rightarrow$ basic $\rightarrow$ polished $\rightarrow$ refined $\rightarrow$ perfect \\
    & Edge Treatment & Discrete & hard, soft, mixed, stylized \\
\bottomrule
\end{tabular}
\end{table}

\section{Reliability of Feature Extraction}
\label{app:feature_extraction}

To validate the extraction accuracy, we measured the agreement between VLM outputs (across three independent runs) and human expert annotations (Gold Standard) for 150 images. The results demonstrate high semantic alignment: the overall mean strict match across all features reaches \(97.00\%\), with the expected pairwise consistency between independent runs averaging \(97.88\%\). The lowest strict match (\(90.00\%\) for \texttt{composition.negative\_space}) stems from the inherent subjectivity of spatial judgments, where even human experts notoriously exhibit high variance.

These results confirm the high reliability of the VLM-based feature extraction. Notably, because we utilized a mid-tier VLM rather than the absolute state-of-the-art (SOTA) model for this process, our reported agreement rates effectively represent a conservative lower bound. Employing a frontier SOTA model would undoubtedly yield even higher accuracy. Furthermore, as foundation models continue to evolve, the precision of this structured extraction will naturally scale, rendering feature extraction stability a non-issue. Moreover, the HRFD closed-loop design ensures inherent robustness to any residual noise: in our cumulative preference tracking, isolated extraction errors regarding subjective features are statistically diluted by persistent signals over multiple rounds, guaranteeing stable convergence.

\begin{table}[!htbp]
\caption{Per-feature VLM extraction reliability across three independent runs (150 images). \textbf{Strict Match} indicates the percentage of images where all three VLM runs unanimously agreed with human expert annotations. \textbf{Mean Pairwise} denotes the average consistency across all three possible pairs of independent runs.}
\label{tab:vlm_reliability}
\centering
\begin{tabular}{@{}llcc@{}}
\toprule
\textbf{Layer} & \textbf{Feature} & \textbf{Strict Match} & \textbf{Mean Pairwise} \\ \midrule
\textbf{L1} 
 & Style: Art Style & 100.00 & 100.00 \\
 & Style: Style Era & 100.00 & 100.00 \\
 & Style: Cultural Style & 99.33 & 99.56 \\
 & Background: Presentation Mode & 99.33 & 99.56 \\
 & Background: Environment & 98.67 & 98.89 \\
 & Emotion: Emotional Type & 98.00 & 98.67 \\ \midrule
\textbf{L2} 
 & Composition: Perspective & 99.33 & 99.56 \\
 & Composition: Visual Flow & 98.67 & 99.11 \\
 & Color: Color Palette & 98.67 & 99.11 \\
 & Color: Light Source Context & 98.00 & 98.44 \\
 & Composition: Layout & 97.33 & 98.00 \\
 & Composition: Framing & 97.33 & 98.00 \\
 & Color: Lighting Design & 96.67 & 97.56 \\
 & Background: Scene Complexity & 95.33 & 96.44 \\
 & Composition: Depth Sense & 94.67 & 96.22 \\
 & Emotion: Emotional Intensity & 92.00 & 94.67 \\
 & Composition: Negative Space & 90.00 & 93.33 \\ \midrule
\textbf{L3} 
 & Background: Background Blur & 99.33 & 99.56 \\
 & Color: Brightness & 99.33 & 99.56 \\
 & Technical: Finish Quality & 98.67 & 99.11 \\
 & Technical: Detail Level & 97.33 & 98.00 \\
 & Color: Temperature & 96.67 & 97.56 \\
 & Color: Saturation & 96.00 & 97.11 \\
 & Technical: Texture Rendering & 95.33 & 96.67 \\
 & Technical: Edge Treatment & 93.33 & 95.33 \\
 & Color: Contrast & 92.67 & 94.89 \\ \midrule
\textbf{Overall} & \textbf{Mean Performance} & \textbf{97.00} & \textbf{97.88} \\ \bottomrule
\end{tabular}%
\end{table}

\section{Design Rationale of the Feature Hierarchy}
\label{app:Design Rationale}

A core challenge in multi-dimensional preference alignment is that when a user has preferences across multiple visual features, it is unlikely for all of them to appear together in a single generated image. Trying to align all features at once therefore leads to slow and unstable convergence. To address this, HRFD organizes features into a three-tier hierarchy and resolves one tier at a time, enforcing a coarse-to-fine convergence strategy that makes it realistic for the generated images to progressively satisfy the user's full set of preferences across rounds.

The three-tier design is justified by two complementary arguments. First, it is consistent with how humans perceive visual content: Global Attributes such as artistic style and visual atmosphere, Spatial and Chromatic Structure such as composition, illumination, and color schemes, and Fine-grained Details such as specific color properties and texture~\cite{hegde2008, navon1977}, directly corresponding to L1$\rightarrow$L2$\rightarrow$L3. Fewer than three tiers would force features of different granularities into the same level, undermining semantic coherence; more than three tiers would make adjacent levels too similar in granularity, reducing the benefit of layering while unnecessarily extending the interaction process. Second, the three-tier structure was not defined a priori but emerged naturally from the Delphi process itself~\cite{dalkey1963}: during the iterative nomination and grouping sessions, the expert photographers independently and consistently organized the candidate features into three distinct granularity levels, providing empirical grounding for the hierarchy from domain experts.

For each identified feature, its candidate values are inherited from existing photography aesthetics literature~\cite{kong2016photo} and further expanded by the five senior photographers during the Delphi process. All candidate values are required to satisfy three criteria: \textbf{perceivability} (users can visually distinguish between different values), \textbf{extractability} (a VLM can reliably identify and differentiate the values, see Appendix \ref{app:feature_extraction} for detailed validation), and \textbf{coverage} (the values collectively cover the full range of real-world variations for that feature). Together, the 
features and their associated values form the complete Hierarchical 
Feature Standard Repository (Table~\ref{tab:table_appendix_features}).

\section{Interactive Procedure Optimization}
\label{app:Interactive_procedure}

\begin{figure}[!htbp]
  \centering
  % [width=\linewidth] 让图片宽度刚好占满整个页面的正文宽度
  \includegraphics[width=\linewidth]{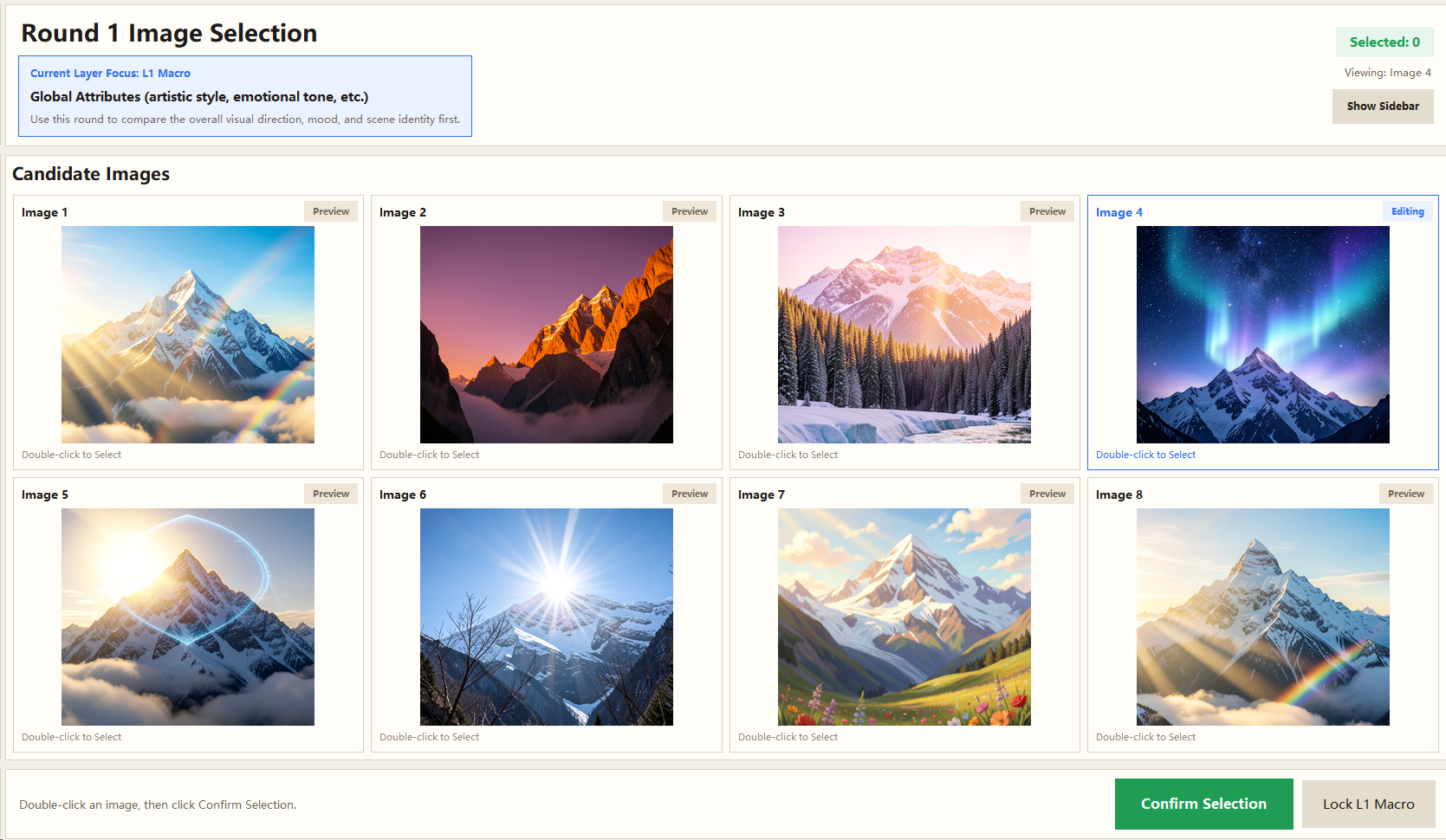}
  \caption{The user interface of our HRFD system.}
  \label{fig:ui_iteration}
\end{figure}

To facilitate efficient convergence, HRFD provides an intuitive interface (Figure~\ref{fig:ui_iteration}) where users progressively refine their intent via binary feedback. This interactive front-end is fundamentally driven by two internal mechanisms: prompt-based locking and tier-level locking.

\textbf{Prompt-Based Locking} is applied before the interaction begins. The system maps the user's initial prompt to the feature repository and identifies any feature values that are already clearly specified. For example, the prompt ``a little girl sleeping indoors'' directly determines the \textit{Environment} feature as \textit{indoor}. These features are immediately locked and written as hard constraints into all subsequent prompt generation steps, completely bypassing preference inference for those dimensions. This ensures that interaction resources are focused entirely on dimensions whose values have not yet been expressed by the user.

\textbf{Tier-Level Locking} governs the transition between feature hierarchies during the interactive phase using user-initiated manual locking and convergence-based auto-locking. Through user-initiated manual locking, if a candidate image precisely satisfies the user's desired feature values for the currently active tier, the user can explicitly select and lock it. This action instantaneously fixes all features within that tier to their corresponding values in the image, concluding the current tier's exploration phase and activating the next tier. 
Concurrently, convergence-based auto-locking guarantees tier convergence without explicit intervention. After each round, the system updates the optimal value for each feature: the candidate value with the highest cumulative Odds Ratio (OR) for discrete features, and the level closest to the preference center $c_{\text{pref}}(f)$ for ordinal features. If a feature's optimal value remains stable for $N$ consecutive rounds, the system autonomously locks it. 
Once all features in the current tier are locked, their values become hard constraints, and the subsequent tier is activated (the threshold $N$ is detailed in Appendix~\ref{app:computational_efficiency}).

In summary, prompt-based locking eliminates predefined constraints before interaction begins, while tier-level locking drives the progressive convergence during the interaction. This ensures exploratory resources are precisely allocated to unresolved dimensions for efficient preference alignment.

\section{Ordinal Preference Analysis}
\label{app:ordinal}

This appendix details the complete mathematical formulations deferred from the main text regarding ordinal features (Sections \ref{sec:ordinal_features} and \ref{sec:gaussian_sampling}). Specifically, it expands upon the variance-based round weight, the weighted statistics within the absolute deviation space, the pooled standard deviation for numerical stability, and the robust sampling variance utilized during the exploitation phase.

\subsection*{Variance-Based Round Weight $w_t(f)$}

The contribution of round $t$ to cumulative preference inference
should reflect how much discriminative information that round
actually provides. If all candidate images in round $t$ share
nearly identical values on feature $f$, user selections carry
negligible signal for inferring whether that feature value is
preferred or not. We capture this intuition via the total variance
of all candidate samples on feature $f$ at round $t$:
\begin{equation}
w_t(f) = \min\!\left(\frac{\sigma_t^2(f)}{0.25},\; 1\right),
\label{eq:weight}
\end{equation}
where $\sigma_t^2(f)$ denotes the variance of feature $f$ across
all candidate images in round $t$, and $0.25$ is the theoretical
maximum variance for any variable bounded in $[0,1]$, serving as
the normalization baseline. Consequently, $w_t(f)$ approaches zero
when all candidates share a near-identical feature value, and
reaches $1$ when candidate diversity is at its maximum, ensuring
that uninformative rounds contribute minimally to the cumulative
preference estimate.

\subsection*{Weighted Mean Deviation $\mu'_g(f)$}

With the cumulative preference center $c_{\text{pref}}(f)$
established (Equation~(5) in the main text), all analysis is
conducted over the absolute deviations of each sample from this
anchor. The variance-weighted mean deviation for group
$g \in \{\mathrm{liked}, \mathrm{disliked}\}$ across all
interaction rounds is defined as:
\begin{equation}
\mu'_g(f) =
  \frac{\displaystyle\sum_{t=1}^{T}
        \sum_{i\in\mathcal{S}_{g,t}}
        w_t(f)\cdot\bigl|v_i^t - c_{\text{pref}}(f)\bigr|}
       {\displaystyle\sum_{t=1}^{T}
        \sum_{i\in\mathcal{S}_{g,t}} w_t(f)},
\label{eq:mu_prime}
\end{equation}
where $v_i^t$ is the feature value of sample $i$ at round $t$,
$\mathcal{S}_{g,t}$ is the set of images in group $g$ at round
$t$, and $w_t(f)$ is the variance-based weight defined in
Equation~(\ref{eq:weight}). A smaller $\mu'_{\mathrm{liked}}(f)$
indicates that liked samples cluster tightly around
$c_{\text{pref}}(f)$, i.e., the user holds a concentrated
preference for that specific feature level.

\subsection*{Weighted Variance $\sigma'^2_g(f)$ and Pooled
Standard Deviation}

The corresponding variance of absolute deviations for group $g$
is computed as:
\begin{equation}
\sigma'^2_g(f) =
  \frac{\displaystyle\sum_{t=1}^{T}
        \sum_{i\in\mathcal{S}_{g,t}}
        w_t(f)\cdot
        \Bigl(\bigl|v_i^t - c_{\text{pref}}(f)\bigr|
              - \mu'_g(f)\Bigr)^2}
       {\displaystyle\sum_{t=1}^{T}
        \sum_{i\in\mathcal{S}_{g,t}} w_t(f)}.
\label{eq:var_prime}
\end{equation}
The pooled standard deviation across the two groups is then:
\begin{equation}
\sigma'_{\mathrm{pooled}}(f)
  = \sqrt{\frac{\sigma'^2_{\mathrm{liked}}(f)
               + \sigma'^2_{\mathrm{disliked}}(f)}{2} + \varepsilon},
\label{eq:pooled_std}
\end{equation}
where $\varepsilon$ is a small constant (set to $10^{-8}$ in our
implementation) that ensures numerical stability when user feedback
is sparse in early interaction rounds, preventing division by zero
in the subsequent Cohen's $d$ computation.

\subsection*{Sampling Variance for Gaussian Exploitation}

As detailed in the main text (Section \ref{sec:gaussian_sampling}), when the cumulative 
effect size $d(f) \ge 0.8$, the HRFD framework shifts to an exploitation 
paradigm, drawing the final continuous feature value $x_f$ from a truncated 
Gaussian distribution: $x_f \sim \mathcal{N}_{[0,1]}(c_{\text{pref}}(f), \sigma_{\mathrm{liked}}^2(f))$. 

To calculate this sampling variance $\sigma_{\mathrm{liked}}^2(f)$, we 
purposefully evaluate the historically liked samples within the 
\textit{original metric space} rather than the absolute deviation space. 
This design choice is critical to prevent variance collapse. Under 
scenarios of symmetrically conflicting feedback (e.g., a user heavily 
favoring both extremes like $0.1$ and $0.9$, but resisting the center), 
calculating variance in the absolute deviation space could cause it to 
artificially degenerate towards zero, thereby trapping the model in a 
narrow sampling point and severely hurting candidate diversity. 

By computing it in the original space, we guarantee a robust and 
representative exploration range around the preference center. 
The rigorous mathematical formulation for this variance is defined as:
\begin{equation}
\sigma^2_{\mathrm{liked}}(f) =
  \frac{\displaystyle\sum_{t=1}^{T}
        \sum_{i\in\mathcal{S}_{\mathrm{liked},t}}
        w_t(f)\cdot \Bigl(v_i^t - c_{\text{pref}}(f)\Bigr)^2}
       {\displaystyle\sum_{t=1}^{T}
        \sum_{i\in\mathcal{S}_{\mathrm{liked},t}} w_t(f)}.
\label{eq:sampling_var}
\end{equation}
Here, $w_t(f)$ is the identical adaptive round weight defined in 
Equation~(\ref{eq:weight}), ensuring that higher-quality feedback 
rounds contribute more significantly to the shape of the generation 
sampling distribution.

\section{Experimental Analysis}
\label{app:experimental_analysis}

\subsection{Qualitative Analysis}
\label{app:Qualitative Analysis}

\begin{figure}[!htbp]
  \centering
  % [width=\linewidth] 确保它横跨两栏占满整行
  \includegraphics[width=\linewidth]{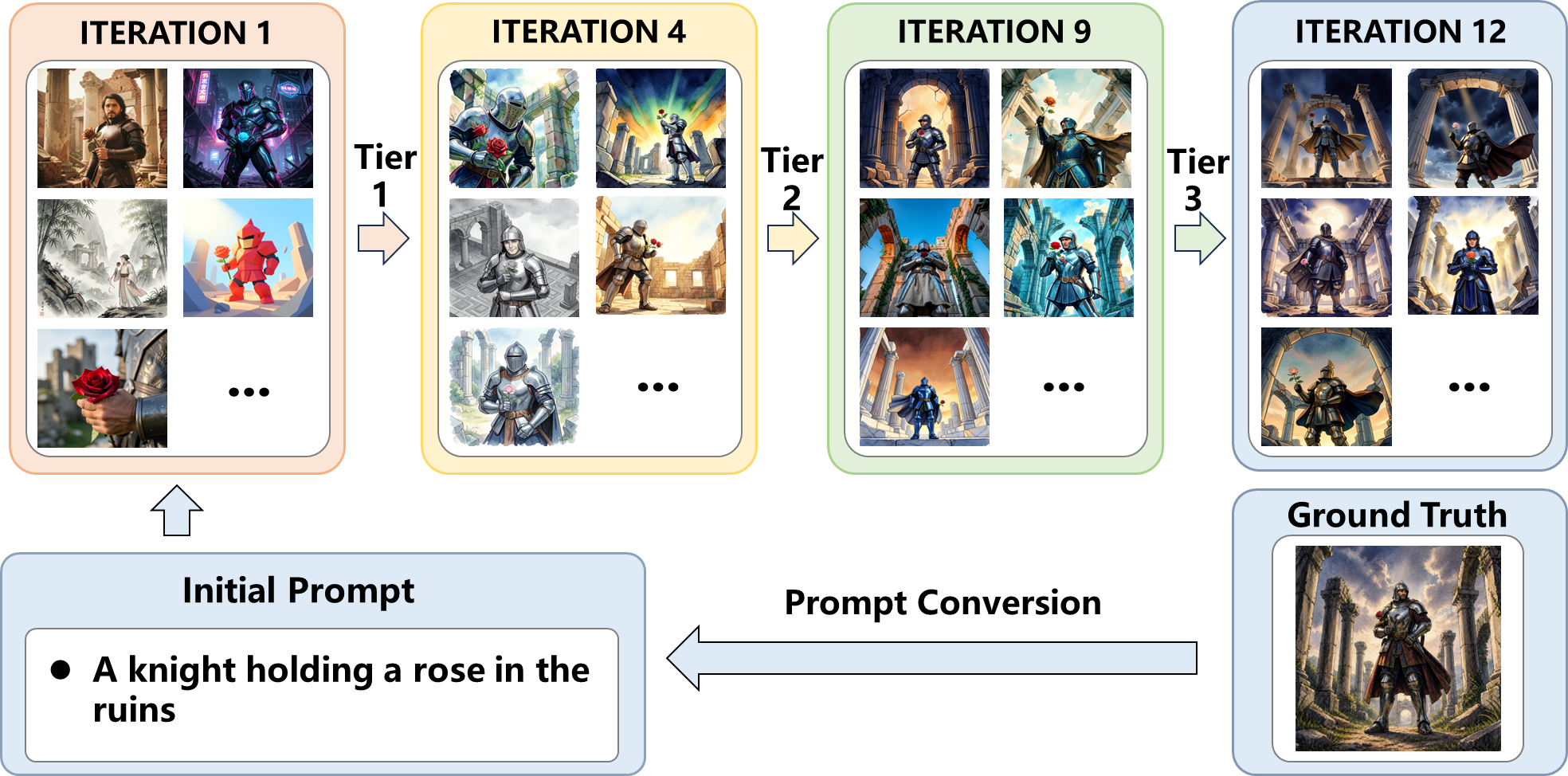} 
  \caption{The progressive convergence process of HRFD.}
  \label{fig:evolution_process}
\end{figure}

To provide a deeper understanding of how HRFD bridges the intention-expression gap, we analyze its progressive convergence mechanism and final visual outcomes. As qualitatively illustrated in Figure~\ref{fig:evolution_process}, we trace an interaction trajectory initialized with a semantically sparse prompt: "A knight holding a rose in the ruins."

\textbf{Tier 1: Global Attribute Convergence.} 
Due to the lack of specific visual constraints in the initial prompt, early generations (Iteration 1) exhibit massive variance, blindly exploring disconnected aesthetic spaces. To address this, HRFD performs preference inference on L1 macro features based on user clicks. By Iteration 4, the framework successfully extracts the latent preference for a watercolor art style, contextualized within a classical and western aesthetic. The environment is firmly anchored as outdoor. This tier establishes the overall style and atmosphere, effectively eliminating stylistic hallucinations for all subsequent generations.

\textbf{Tier 2: Spatial and Chromatic Structure Alignment.} 
With the global style securely locked, HRFD transitions to optimizing L2 meso features, focusing on spatial organization and chromatic structure. During this phase, user feedback is utilized to progressively resolve spatial and colorimetric ambiguities. By Iteration 9, HRFD robustly converges on a low\_angle perspective, wide framing, and a strictly centered layout characterized by deep spatial depth. Concurrently, it establishes the chromatic foundation, utilizing natural yet dramatic illumination to shape and enhance a complementary color palette. At this stage, both the spatial and chromatic structures of the target intent are fully formulated.

\textbf{Tier 3: Fine-Grained Detail Calibration.} 
In the final refinement phase, HRFD operates on the L3 micro tier to fine-tune specific visual details. Building upon the chromatic structure established in Tier 2, it dictates the exact color manifestation, explicitly converging on a warm color temperature, moderate saturation, and dim brightness. Furthermore, HRFD refines micro-details, enforcing refined texture rendering, a high level of detail, and a sharp background under extreme contrast. By Iteration 12, the generated candidate closely aligns with the visual intent of the Ground Truth.

\begin{figure}[!htbp]
  \centering
  \includegraphics[width=\linewidth]{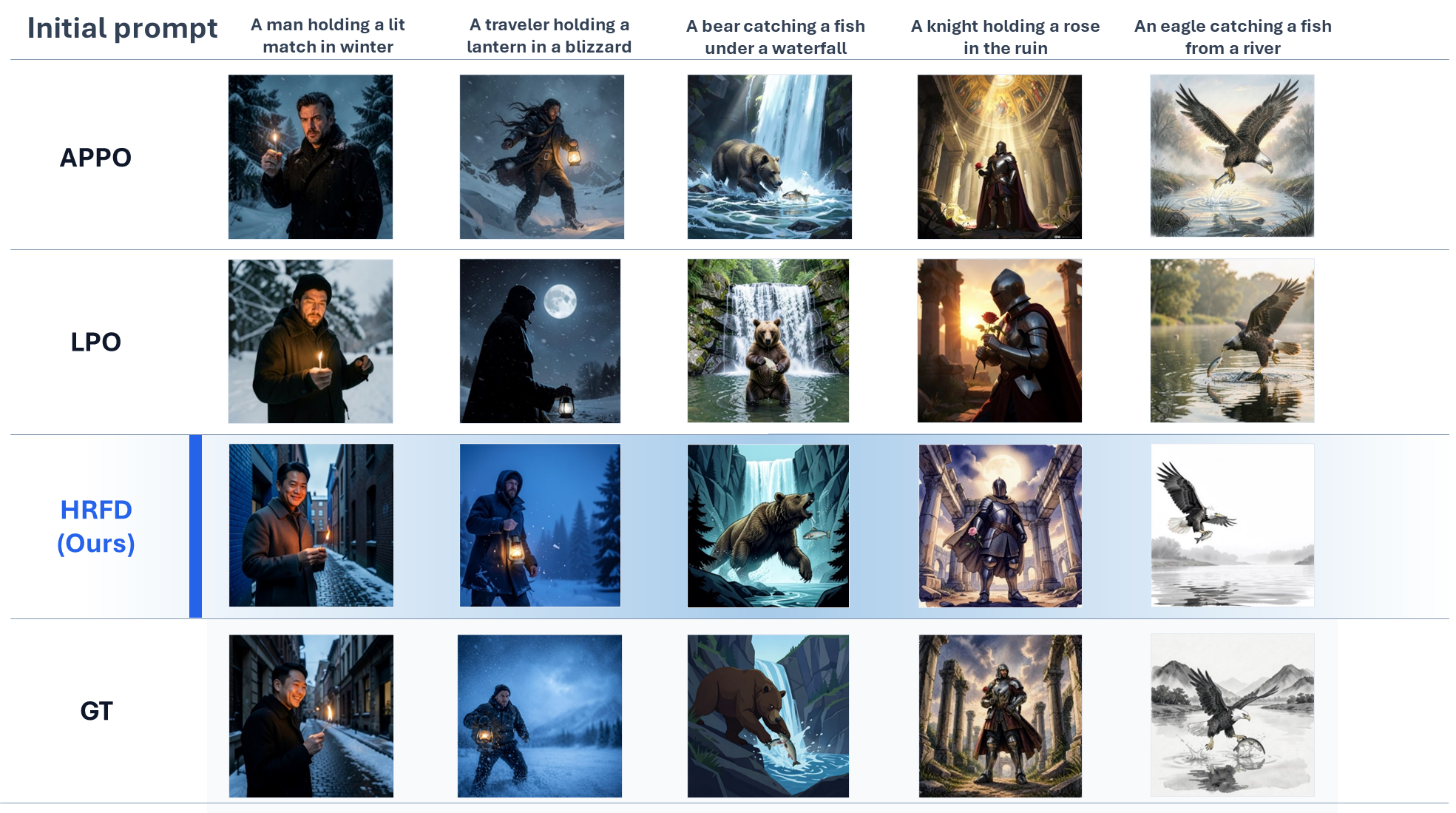} 
  \caption{Qualitative comparison of representative generated images across different methods in the Objective Preference Alignment scenario. Compared to LPO and APPO, our HRFD framework successfully captures multi-dimensional user preferences, generating results that most closely align with the Ground Truth (GT).}
  \label{fig:qualitative_comparison}
\end{figure}

In summary, this hierarchical progression demonstrates how HRFD systematically achieves coarse-to-fine preference alignment. By strictly decoupling macro, meso, and micro features, the framework simplifies the complex multi-dimensional alignment process, efficiently translating abstract mental intents into precise visual mappings.
The superiority of the HRFD is visually validated in 
Figure~\ref{fig:qualitative_comparison}, which compares HRFD against LPO and APPO in diverse representative scenarios.

\subsection{Ablation Analysis}
\label{app:ablation_analysis}
The removal of the hierarchical feature architecture (\textit{w/o Hierarchy}) inflicts the most severe degradation across alignment metrics (CLIP: 0.7553, SSIM: 0.1491) and efficiency (14.63 rounds). The failure stems from binary signal contamination. In multi-dimensional alignment tasks, users typically select an image during iterations because a few currently focused features meet their needs. Updating all dimensions at once incorrectly reinforces unwanted features because they are bundled within the chosen image. 
Empirical feedback explicitly corroborates this: participants reported that if attempting to align all desired features during the iterative process, they had to constantly alternate their focus among different features across rounds to ensure these desired attributes would appear in subsequent iterations. This shifting focus induced mutual signal contamination that prevented simultaneous convergence. This variant only succeeded when users abandoned some features and settled for aligning a few dominant features.
This phenomenon directly underscores the necessity of our hierarchical design, which strategically isolates feature sets to guarantee that preference updates strictly reflect the user's current focus without affecting other dimensions. Therefore, while this variant might suffice for partial adjustments, it collapses under comprehensive multi-dimensional demands.

Disabling the global preference memory (\textit{w/o History}) substantially degrades alignment across all metrics (CLIP: 0.8196, LPIPS: 0.7610, SSIM: 0.1903) and convergence efficiency (14.20 rounds). This failure stems from \textit{preference amnesia}. Without cross-round accumulation, inference relies exclusively on the immediate preceding feedback, completely discarding previously established preferences. Consequently, generated images only reflect the most recently expressed intent, leaving feature coverage perpetually incomplete. This variant therefore fails to build the sustained preference signals required for multi-dimensional alignment, confirming that cumulative preference inference is strictly indispensable.

Finally, eliminating the dynamic weighting mechanism (\textit{w/o Weighting}) disrupts convergence efficiency (12.93 rounds), which subsequently degrades the final alignment accuracy (CLIP: 0.8334, LPIPS: 0.7346, SSIM: 0.2186). This performance drop arises because treating all interaction rounds equally makes this variant highly vulnerable to low-information-quality data. For instance, if almost all candidate images in a round happen to share a specific feature value, the user's selections carry negligible discriminative power for inferring whether that feature value is actually preferred. Without adaptive weighting, this variant treats such rounds equally with highly discriminative ones. Consequently, if a user selects only a few images in such a round, the variant incorrectly interprets the unselected majority as evidence of a negative preference toward that feature value, corrupting the cumulative preference estimate and misdirecting subsequent generation away from the user's true intent. 
By dynamically suppressing the contribution of uninformative rounds, the full HRFD framework prevents such misleading signals from contaminating the preference state, ensuring stable and faithful convergence.

\section{Computational Efficiency and Implementation Details}
\label{app:computational_efficiency}

Table~\ref{tab:hyperparameters} summarizes the key hyperparameters used throughout all experiments. These values were fixed across both Scenario~I and Scenario~II to ensure a fair and consistent comparison. 

\begin{table}[!htbp]
\caption{Key Hyperparameters for the HRFD Framework}
\label{tab:hyperparameters}
\centering
% 将行间距倍数从 1.2 提高到 1.5，让表格看起来更舒展
\renewcommand{\arraystretch}{1.5} 
\setlength{\tabcolsep}{6pt} 
\begin{tabular}{@{}llc@{}}
\toprule
\textbf{Category} & \textbf{Parameter} & \textbf{Value} \\
\midrule
\multirow{3}{*}{\textbf{Interaction}} 
& Max Interaction Rounds ($T_{max}$)& 15 \\
& Candidate Images per Round ($N$) & 8 \\  
& Auto-locking Threshold ($N_{\text{lock}}$) & 4 \\  
\midrule
\multirow{4}{*}{\textbf{Generation}} 
& Generation Backend & FLUX.2-Klein (Distilled)\\
& Image Resolution & $512 \times 512$ \\
& Sampling Steps & 4 \\
& Guidance Scale & 1.0 \\
\bottomrule
\end{tabular}
\end{table}

Table~\ref{tab:latency} details the latency of HRFD per interaction round (generating $N=8$ candidate images). The framework comprises four stages: Feature Extraction, Preference Inference, Prompt Generation, and Image Rendering. 

Due to hardware constraints, the VLM-based feature extraction and prompt generation are executed via parallelized external API calls, averaging 3.68s (range: 2.92--5.17s) and 7.25s (range: 6.76--7.65s), respectively. Crucially, our closed-form mathematical preference inference completely avoids neural network overhead, computing in under 1 ms. Finally, the local image rendering is executed serially, averaging 2.61s per image ($\sim$20.88s total). 

While the current end-to-end latency averages 31.81s per round, this is strictly a software engineering bottleneck rather than an algorithmic flaw. The current latency is primarily an artifact of network-based API calls and serial rendering on limited hardware. Moving to edge-deployed VLMs and batched parallel tensor operations on dedicated GPUs will deterministically reduce this to sub-5 seconds, ensuring fluid real-time interaction. Together, these optimizations fully unlock HRFD's potential for real-time interactive applications.

\begin{table}[!htbp]
  \caption{Latency decomposition of the HRFD framework per interaction round.}
  \label{tab:latency}
  \centering
  \renewcommand{\arraystretch}{1.2}
  \begin{tabular}{@{}lccc@{}}
      \toprule
      \textbf{Stage} & \textbf{Avg. Latency} & \textbf{Execution Mode} & \textbf{Computing Backend} \\
      \midrule
      Feature Extraction & 3.68 s & Parallel API & VLM \\
      Preference Inference & $<$1 ms & Local Compute & CPU (Algorithm) \\
      Prompt Generation & 7.25 s & Parallel API & VLM \\
      Image Rendering & 20.88 s & Serial Local & Flux.2-Klein \\
      \midrule
      \textbf{Total Round Latency} & \textbf{31.81 s} & \textbf{Mixed} & \textbf{Current Config.} \\
      \bottomrule
  \end{tabular}
\end{table}

\section{Meta-Prompts used in HRFD}
\label{app:meta_prompts}

In this section, we present the comprehensive meta-prompts utilized in our HRFD framework. This includes the prompt for autonomous Ground Truth (GT) generation to ensure objective evaluation, as well as the prompts driving our core interactive pipeline: the structured feature extraction phase and the subsequent prompt generation phase.

\subsection{Ground Truth Generation Meta-Prompt}
\label{app:gt_generation_metaprompt}

To ensure objective evaluation in Scenario I, we employ a Foundation Model to autonomously synthesize highly detailed prompts for Ground Truth (GT) image generation. This automated approach mitigates human bias and ensures the GT images exhibit diverse and rich visual attributes across dimensions such as style, composition, and lighting. The exact meta-prompt is provided below.

\Needspace{18\baselineskip}
\begin{lstlisting}[style=promptstyle]
You are an expert prompt engineer for text-to-image generative models.

## Your task is:
Generate {num_cases} distinct, highly detailed image-generation prompts to serve as Ground Truth (GT) references.

## Requirements:
- Subject Diversity: Cover a diverse range of core subjects (e.g., human portraits, animal wildlife, natural landscapes, and architectural scenes).
- Rich Visual Details: Each prompt must be comprehensively descriptive. Explicitly specify visual attributes across multiple dimensions, such as artistic style, color palette, lighting design, camera perspective, and composition.
- Visual Explicitness: Avoid vague aesthetic terms. Translate target features into observable physical descriptions (e.g., instead of "warm temperature," specify "bathed in warm, golden-hour sunlight casting long shadows").

## Output format:
Output only valid JSON strictly following this schema:
{
  "gt_cases": [
    {
      "case_id": 1,
      "base_subject": "A brief description of the core subject (e.g., 'a sleeping cat')",
      "detailed_prompt": "The final, highly detailed prompt seamlessly weaving the base subject, style, lighting, and composition into a comprehensive descriptive paragraph."
    },
    ...
  ]
}
\end{lstlisting}

\subsection{Feature Extraction Meta-Prompt}
\label{app:stage1_metaprompt}

For reproducibility, we report the meta-prompt used to guide layer-specific structured feature extraction.

\Needspace{18\baselineskip}
\begin{lstlisting}[style=promptstyle]
You are an expert image analyst. Analyze {num_images} images and extract structured visual features.

## Output only valid JSON with an "images" array.

## The feature repository is organized into 3 layers:
- L1 (Macro): style.art_style, style.style_era, style.cultural_style, emotion.emotional_type, background.environment
- L2 (Meso): color.color_palette, color.light_source_context, color.lighting_design, composition.framing, composition.perspective, composition.layout, composition.visual_flow, composition.depth_sense, composition.negative_space, background.presentation_mode, background.scene_complexity, emotion.emotional_intensity
- L3 (Micro): color.temperature, color.saturation, color.brightness, color.contrast, background.background_blur, technical.detail_level, technical.texture_rendering, technical.edge_treatment, technical.finish_quality

## Current extraction target:
- Active layer: {active_layer}
- Extract only the features belonging to the active layer.
- Do not infer or output features from other layers.

## Use only the predefined vocabulary below:

## Style
- art_style: realistic | illustration | flat_illustration | cartoon | anime | 3d_render | pixel_art | watercolor | oil_painting | sketch | line_art | minimalist | surreal | collage | mixed_media
- style_era: classical | vintage | modern | futuristic
- cultural_style: asian | western | african | middle_eastern | mixed

## Color
- temperature: cold | cool | neutral | warm | hot
- color_palette: monochrome | limited_palette | analogous | complementary | triadic | varied
- saturation: desaturated | muted | moderate | vibrant | oversaturated
- brightness: dark | dim | medium | bright | very_bright
- contrast: low | medium_low | medium | high | extreme
- light_source_context: natural | studio | neon | ambient
- lighting_design: flat | soft_diffused | backlit | rim | dramatic

## Composition
- framing: extreme_closeup | closeup | medium | wide | extreme_wide
- perspective: flat | eye_level | low_angle | high_angle | birds_eye | isometric
- layout: centered | rule_of_thirds | symmetrical | asymmetrical | diagonal | radial | scattered
- depth_sense: flat | shallow | moderate | deep | extreme_deep
- negative_space: cramped | minimal | balanced | abundant | dominant
- visual_flow: static | dynamic

## Background
- presentation_mode: solid | gradient | patterned | abstract_graphic | scenic
- scene_complexity: minimal | simple | moderate | detailed | complex
- environment: indoor | outdoor | studio | nature | urban | abstract | fantasy | mixed
- background_blur: sharp | slightly_blurred | heavily_blurred

## Emotion
- emotional_type: serene | joyful | romantic | melancholic | mysterious | tense | angry | playful | ominous
- emotional_intensity: subtle | mild | moderate | strong | overwhelming

## Technical
- detail_level: low | medium_low | medium | high | ultra
- texture_rendering: flat | subtle | detailed | hyper_realistic
- edge_treatment: hard | soft | mixed | stylized
- finish_quality: rough | basic | polished | refined | perfect

## Inference principles:
- Treat this as a closed-vocabulary visual classification task rather than open-ended description.
- For each field, compare all candidate labels within that field and choose the single label best supported by the visible image.
- Base decisions on stable global visual evidence rather than semantic guesswork, narrative association, or a single local cue.
- When several labels appear plausible, select the visually dominant one and apply the same decision standard consistently across all images.
- Only output features from the active layer.

## Output schema:
{
  "images": [
    {
      "image_id": 1,
      "filename": "xxx.png",
      "style": {
        "art_style": "...",
        "style_era": "...",
        "cultural_style": "..."
      },
      ...
    },
    ...
  ]
}
\end{lstlisting}

\subsection{Prompt Generation Meta-Prompt}
\label{app:stage3_metaprompt}

For reproducibility, we report the meta-prompt used to convert selected feature sets into image-generation prompts.

\Needspace{18\baselineskip}
\begin{lstlisting}[style=promptstyle]
You are an expert prompt engineer for image generation.

## You will be given:
- the user's original request,
- {num_prompts} sets of suggested visual features, and
- optional feature-priority information indicating which features belong to determined tiers and which belong to the current tier.

## Your task is:
To convert each feature set into coherent image-generation prompts.

## Requirements:
- The user's original request is the highest-priority constraint and must be preserved in every prompt.
- The subject, action, and core scene intent specified by the user must remain unchanged.
- Suggested features should be incorporated as fully as possible, provided that they do not alter or contradict the user's original request.

## Feature realization:
- Express features through concrete visual evidence rather than raw feature labels.
- Do not simply restate terms such as "warm," "high_angle," "rule_of_thirds," or "dramatic lighting."
- Instead, translate features into observable image content, for example:
  composition through crop, camera position, subject placement, spacing, and directional structure;
  color through specific hues and palette relationships;
  lighting through source, direction, shadow behavior, and illumination pattern;
  background through recognizable scene elements;
  emotion through facial expression, body language, atmosphere, color, and lighting.

## Coherence:
- Each prompt should read as a unified scene description rather than a list of attributes.
- If features conflict, preserve the user's original request first, then preserve features from determined tiers, and only omit lower-priority features from the current tier when necessary.

## Tier priority:
- When feature-priority information is provided, treat features from determined tiers as fixed constraints.
- Express features from determined tiers earlier and more prominently in the prompt.
- Add features from the current tier only after the determined-tier features have already been clearly established.
- Features from the current tier must not weaken or override features from determined tiers.
- Features from not-yet-considered tiers should not be introduced into the prompt.

## Writing guidance:
- Prompts should typically be 30-80 words, though slightly longer descriptions are acceptable when required for clear grounding.
- Use concise, specific, and visually explicit language.
- Every applied feature should be identifiable from the generated image itself.

## Output format:
{
  "prompts": [
    {
      "image_number": 1,
      "concept": "...",
      "prompt": "...",
      "applied_features": ["...", "..."]
    },
    ...
  ]
}
\end{lstlisting}

\section{Detailed Statistical Results for Subjective Evaluation}
\label{app:statistical_results}

To rigorously compare HRFD with the APPO baseline, we applied a Wilcoxon signed-rank test to the subjective metrics gathered during the user study (Interaction Rounds, NASA-TLX, and CSI). This section details the statistical results, specifically reporting the test statistics ($W$) and their corresponding $p$-values. Note that the NASA-TLX metrics are measured on a 0–100 scale, whereas the Creativity Support Index (CSI) utilizes a 0–10 scale. Ultimately, these results demonstrate that HRFD delivers statistically significant improvements across key efficiency, alignment, and satisfaction dimensions.
\begin{table}[!htbp]
\centering
\caption{Wilcoxon Signed-Rank Test Results for Interaction Rounds, NASA-TLX, and CSI Metrics}
\label{tab:statistical_significance}
\renewcommand{\arraystretch}{1.2}
\begin{tabular}{lcc}
\toprule
\textbf{Metric} & \textbf{Test Statistic ($W$)} & \textbf{$p$-value} \\
\midrule
\rowcolor{mygray!30} \multicolumn{3}{l}{\textit{Efficiency Metric}} \\
Interaction Rounds $\downarrow$ & 0.00 & $0.0073^{**}$ \\
\midrule
\rowcolor{mygray!30} \multicolumn{3}{l}{\textit{NASA-TLX Metrics (0--100)}} \\
Mental Demand $\downarrow$ & 23.00 & $0.0349^{*}$ \\
Physical Demand $\downarrow$ & 58.50 & $0.9320$ \\
Temporal Demand $\downarrow$ & 19.50 & $0.0213^{*}$ \\
Performance Dissatisfaction $\downarrow$ & 7.50 & $0.0029^{**}$ \\
Effort $\downarrow$ & 23.00 & $0.0354^{*}$ \\
Frustration $\downarrow$ & 6.00 & $0.0022^{**}$ \\
\midrule
\rowcolor{mygray!30} \multicolumn{3}{l}{\textit{CSI Metrics (0--10)}} \\
Enjoyment $\uparrow$ & 15.00 & $0.0152^{*}$ \\
Exploration $\uparrow$ & 4.00 & $0.0014^{**}$ \\
Expressiveness $\uparrow$ & 10.00 & $0.0043^{**}$ \\
Immersion $\uparrow$ & 25.00 & $0.7907$ \\
Results Worth Effort $\uparrow$ & 17.00 & $0.0137^{*}$ \\
\bottomrule
\multicolumn{3}{l}{\footnotesize $^{*}p < 0.05, ^{**}p < 0.01$; $\downarrow$ lower is better, $\uparrow$ higher is better.}
\end{tabular}
\end{table}

\FloatBarrier
\newpage
\section*{NeurIPS Paper Checklist}

\begin{enumerate}

\item {\bf Claims}
    \item[] Question: Do the main claims made in the abstract and introduction accurately reflect the paper's contributions and scope?
    \item[] Answer: \answerYes{} % Replace by \answerYes{}, \answerNo{}, or \answerNA{}.
    \item[] Justification: The abstract and introduction clearly state the contributions of the HRFD framework, which are fully supported by the quantitative and qualitative experimental results.

\item {\bf Limitations}
    \item[] Question: Does the paper discuss the limitations of the work performed by the authors?
    \item[] Answer: \answerYes{} % Replace by \answerYes{}, \answerNo{}, or \answerNA{}.
    \item[] Justification: We have explicitly discussed the limitations of our work in the "Conclusion and Limitations" section, specifically addressing the Out-of-Vocabulary (OOV) bottleneck caused by the static feature repository.

\item {\bf Theory assumptions and proofs}
    \item[] Question: For each theoretical result, does the paper provide the full set of assumptions and a complete (and correct) proof?
    \item[] Answer: \answerNA{} % Replace by \answerYes{}, \answerNo{}, or \answerNA{}.
    \item[] Justification: This paper proposes an interactive, training-free framework for text-to-image generation and does not introduce new theoretical theorems or mathematical proofs.

    \item {\bf Experimental result reproducibility}
    \item[] Question: Does the paper fully disclose all the information needed to reproduce the main experimental results of the paper to the extent that it affects the main claims and/or conclusions of the paper (regardless of whether the code and data are provided or not)?
    \item[] Answer: \answerYes{} % Replace by \answerYes{}, \answerNo{}, or \answerNA{}.
    \item[] Justification: We provide all necessary details for reproducibility, including key hyperparameters in Appendix G (Table 5), the complete feature repository in Appendix A, and all meta-prompts in Appendix H.

\item {\bf Open access to data and code}
    \item[] Question: Does the paper provide open access to the data and code, with sufficient instructions to faithfully reproduce the main experimental results, as described in supplemental material?
    \item[] Answer: \answerNo{} % Replace by \answerYes{}, \answerNo{}, or \answerNA{}.
    \item[] Justification: While we do not provide a direct URL to a code repository in the main text, the exact algorithms, inference equations, and full meta-prompts are thoroughly documented in the paper and appendices to ensure verifiability.

\item {\bf Experimental setting/details}
    \item[] Question: Does the paper specify all the training and test details (e.g., data splits, hyperparameters, how they were chosen, type of optimizer) necessary to understand the results?
    \item[] Answer: \answerYes{} % Replace by \answerYes{}, \answerNo{}, or \answerNA{}.
    \item[] Justification: The experimental setup, including the 15 human participants, the specific backend model (FLUX.2-Klein), and interaction parameters, are detailed in Section 4.1 and Appendix G (Table 5).

\item {\bf Experiment statistical significance}
    \item[] Question: Does the paper report error bars suitably and correctly defined or other appropriate information about the statistical significance of the experiments?
    \item[] Answer: \answerYes{} % Replace by \answerYes{}, \answerNo{}, or \answerNA{}.
    \item[] Justification: Quantitative results in Table 1 report both the Mean and Standard Deviation (±SD).Additionally, statistical significance for subjective metrics is evaluated using the Wilcoxon signed-rank test, with p-values reported in Figure 3 and Appendix I.

\item {\bf Experiments compute resources}
    \item[] Question: For each experiment, does the paper provide sufficient information on the computer resources (type of compute workers, memory, time of execution) needed to reproduce the experiments?
    \item[] Answer: \answerYes{} % Replace by \answerYes{}, \answerNo{}, or \answerNA{}.
    \item[] Justification: We disclose the computing hardware (Intel i7-12700KF, 22GB RTX 2080 Ti GPU) in Section 4.1 and provide a detailed latency decomposition per interaction round in Appendix G.
    
\item {\bf Code of ethics}
    \item[] Question: Does the research conducted in the paper conform, in every respect, with the NeurIPS Code of Ethics \url{https://neurips.cc/public/EthicsGuidelines}?
    \item[] Answer: \answerYes{} % Replace by \answerYes{}, \answerNo{}, or \answerNA{}.
    \item[] Justification: The research conforms to the NeurIPS Code of Ethics. The human-in-the-loop study involves benign image preference feedback without malicious use cases or privacy violations.

\item {\bf Broader impacts}
    \item[] Question: Does the paper discuss both potential positive societal impacts and negative societal impacts of the work performed?
    \item[] Answer: \answerNo{} % Replace by \answerYes{}, \answerNo{}, or \answerNA{}.
    \item[] Justification: The paper focuses on a foundational framework to bridge the intention-expression gap and does not include a dedicated section on broader societal impacts, as it does not introduce new architectures that fundamentally alter the safety of existing models.
    
\item {\bf Safeguards}
    \item[] Question: Does the paper describe safeguards that have been put in place for responsible release of data or models that have a high risk for misuse (e.g., pre-trained language models, image generators, or scraped datasets)?
    \item[] Answer: \answerNA{} % Replace by \answerYes{}, \answerNo{}, or \answerNA{}.
    \item[] Justification: Our method is a purely training-free inference framework. We do not release new pre-trained foundation models or scraped internet datasets that pose a high risk of misuse.

\item {\bf Licenses for existing assets}
    \item[] Question: Are the creators or original owners of assets (e.g., code, data, models), used in the paper, properly credited and are the license and terms of use explicitly mentioned and properly respected?
    \item[] Answer: \answerYes{} % Replace by \answerYes{}, \answerNo{}, or \answerNA{}.
    \item[] Justification: We properly cite all existing models and tools used, including FLUX, the LAION aesthetic predictor, and standard evaluation metrics (NASA-TLX, CSI).

\item {\bf New assets}
    \item[] Question: Are new assets introduced in the paper well documented and is the documentation provided alongside the assets?
    \item[] Answer: \answerNA{} % Replace by \answerYes{}, \answerNo{}, or \answerNA{}.
    \item[] Justification: The paper proposes an interactive methodology and does not introduce or release new datasets or pre-trained models as assets.

\item {\bf Crowdsourcing and research with human subjects}
    \item[] Question: For crowdsourcing experiments and research with human subjects, does the paper include the full text of instructions given to participants and screenshots, if applicable, as well as details about compensation (if any)? 
    \item[] Answer: \answerYes{} % Replace by \answerYes{}, \answerNo{}, or \answerNA{}.
    \item[] Justification: Section 4.1 outlines the experimental setup with 15 human participants, explicitly describing the standardized training and the four operational principles provided to them.

\item {\bf Institutional review board (IRB) approvals or equivalent for research with human subjects}
    \item[] Question: Does the paper describe potential risks incurred by study participants, whether such risks were disclosed to the subjects, and whether Institutional Review Board (IRB) approvals (or an equivalent approval/review based on the requirements of your country or institution) were obtained?
    \item[] Answer: \answerNA{} % Replace by \answerYes{}, \answerNo{}, or \answerNA{}.
    \item[] Justification: The user study consists exclusively of participants clicking on synthetic generated images to indicate aesthetic preferences, which does not incur physical or psychological risks requiring IRB approval under our institution's guidelines.

\item {\bf Declaration of LLM usage}
    \item[] Question: Does the paper describe the usage of LLMs if it is an important, original, or non-standard component of the core methods in this research? Note that if the LLM is used only for writing, editing, or formatting purposes and does \emph{not} impact the core methodology, scientific rigor, or originality of the research, declaration is not required.
    %this research? 
    \item[] Answer: \answerYes{} % Replace by \answerYes{}, \answerNo{}, or \answerNA{}.
    \item[] Justification: Large Language Models/Vision-Language Models (e.g., Llama-4-Maverick-17B) are explicitly stated as core components of the HRFD framework for feature extraction and prompt generation, with complete usage details and meta-prompts provided in Section 3 and Appendix H.

\end{enumerate}

\end{document}